\documentclass{article}

\usepackage[nonatbib,final]{neurips_2024}

\usepackage[utf8]{inputenc} 
\usepackage[T1]{fontenc}
\usepackage[numbers]{natbib}
\usepackage{url}            
\usepackage{booktabs}       
\usepackage{amsfonts}       
\usepackage{nicefrac}       
\usepackage{microtype}      
\usepackage{xcolor}         
\usepackage{graphicx}
\usepackage{amsmath}
\usepackage{wrapfig}
\usepackage{adjustbox} 
\usepackage{multirow}
\usepackage{paralist}
\usepackage{CJKutf8}
\usepackage{subcaption}
\usepackage{colortbl}
\usepackage{tgpagella} 
\usepackage{mathpazo}  
\usepackage{XCharter} 
\usepackage[textsize=tiny]{todonotes}
\usepackage{arydshln}
\usepackage[most]{tcolorbox}
\usepackage{enumitem}
\definecolor{newgreen}{rgb}{0.7, 0.9, 0.7}
\definecolor{newblue}{rgb}{0.85, 0.85, 0.9}

\newcommand{\modelname}{\texttt{ECP}}
\definecolor{tkcolor}{RGB}{224,223,255}
\definecolor{shallowred}{RGB}{242,204,208}
\newtcolorbox{takeaways}[1][]{
	width=\columnwidth,
	colback = tkcolor, 
	colframe = tkcolor, 
	boxsep=0pt,left=10pt,right=10pt,top=5pt,bottom=5pt,
	fontupper=\linespread{0.9}\selectfont,
	title=#1}

\newtcolorbox{limitation}[1][]{
	width=\columnwidth,
	colback = shallowred, 
	colframe = shallowred, 
	boxsep=0pt,left=10pt,right=10pt,top=5pt,bottom=5pt,
	fontupper=\linespread{0.9}\selectfont,
	title=#1}
\definecolor{shallowgray}{RGB}{237,240,246}
\newtcolorbox{prompt}[1][]{
	width=\columnwidth,
	colback = shallowgray, 
	colframe = shallowgray, 
	boxsep=0pt,left=10pt,right=10pt,top=5pt,bottom=5pt,
	fontupper=\linespread{0.9}\selectfont,
	title=#1}
\usepackage{hyperref}       
\definecolor{darkblue}{rgb}{0, 0.40, 0.75}
\hypersetup{colorlinks=true, citecolor=darkblue, linkcolor=darkblue, urlcolor=darkblue}

\title{\fontsize{15pt}{15pt}\selectfont Electronic Circuit Principles of Large Language Models}

\author{
	Qiguang Chen$^{\dagger}$ \quad Libo Qin$^{\ddagger}$\thanks{Corresponding Author} \quad Jinhao Liu$^{\dagger}$ \quad Dengyun Peng$^{\dagger}$ \quad Jiaqi Wang$^\diamondsuit$ \\
	\textbf{Mengkang Hu$^\clubsuit$ \quad Zhi Chen$^\spadesuit$ \quad Wanxiang Che$^{\dagger}$\footnotemark[1] \quad Ting Liu$^{\dagger}$ } \\
	$^{\dagger}$ Harbin Institute of Technology \quad $^{\ddagger}$ Central South University \\
	$^\diamondsuit$ The Chinese University of Hong Kong \quad
	$^\clubsuit$ The University of Hong Kong \\
	$^\spadesuit$ ByteDance Seed (China) \\
	\texttt{\{qgchen,car\}@ir.hit.edu.cn},  \texttt{lbqin@csu.edu.cn} \\
}

\begin{document}
	
	\maketitle
	\vspace{-18pt}
	\begin{figure}[!h]
		\centering
		\includegraphics[width=0.9\textwidth]{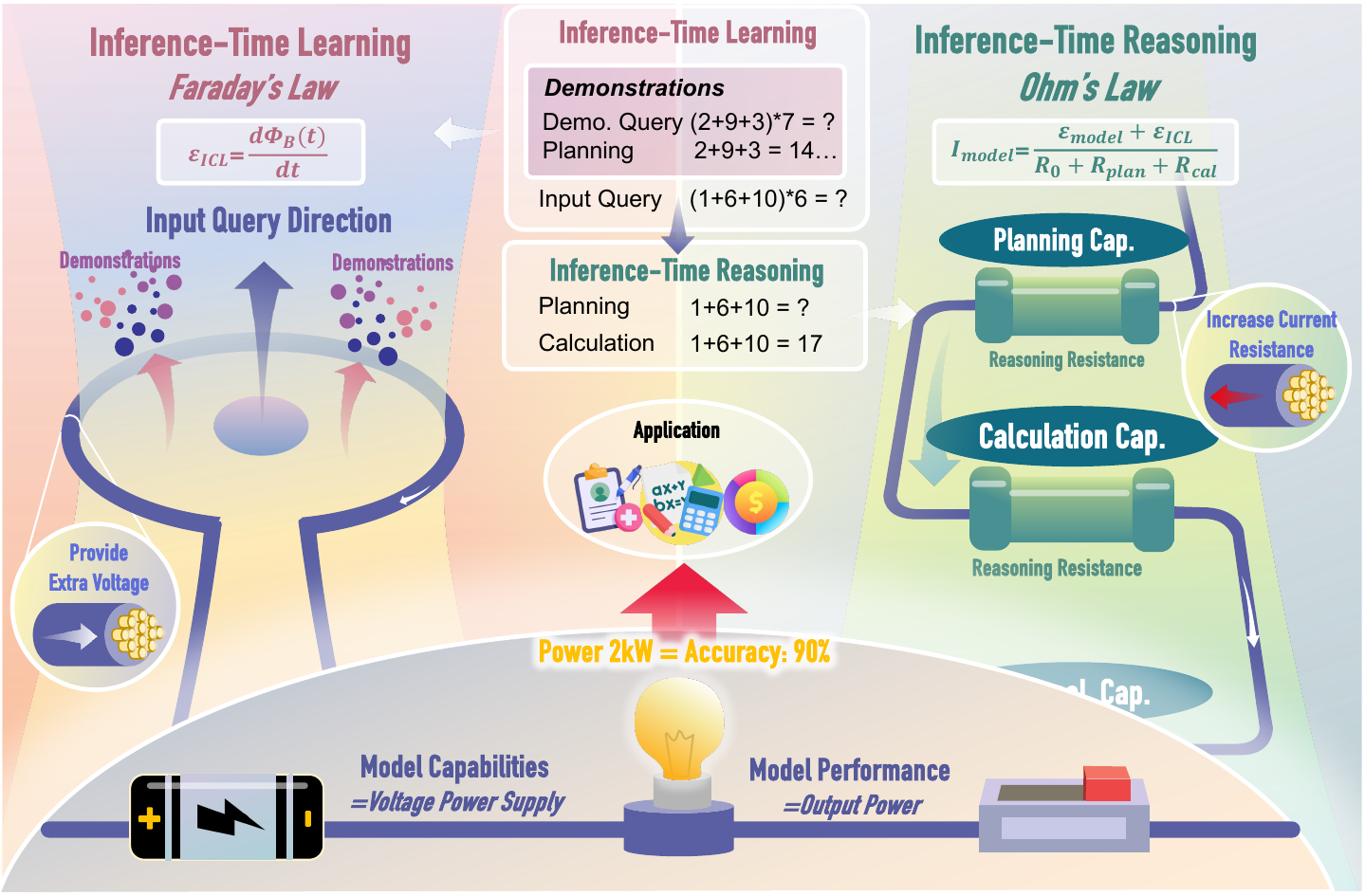}
	\end{figure}
	\begin{abstract}
		Large language models (LLMs) such as DeepSeek-R1 have achieved remarkable performance across diverse reasoning tasks. To uncover the principles that govern their behaviour, we introduce the Electronic Circuit Principles (ECP), which maps inference-time learning (ITL) onto a semantic electromotive force and inference-time reasoning (ITR) onto a resistive network governed by Ohm’s and Faraday’s laws. This circuit-based modelling yields closed-form predictions of task performance and reveals how modular prompt components interact to shape accuracy. We validated ECP on 70,000 samples spanning 350 reasoning tasks and 9 advanced LLMs, observing a about 60\% improvement in Pearson correlation relative to the conventional inference-time scaling law. Moreover, ECP explains the efficacy of 15 established prompting strategies and directs the development of new modular interventions that exceed the median score of the top 80\% of participants in both the International Olympiad in Informatics and the International Mathematical Olympiad. By grounding LLM reasoning in electronic-circuit principles, ECP provides a rigorous framework for predicting performance and optimising modular components.

	\end{abstract}

\begin{figure*}[!t]
    \centering
    \includegraphics[width=0.98\textwidth]{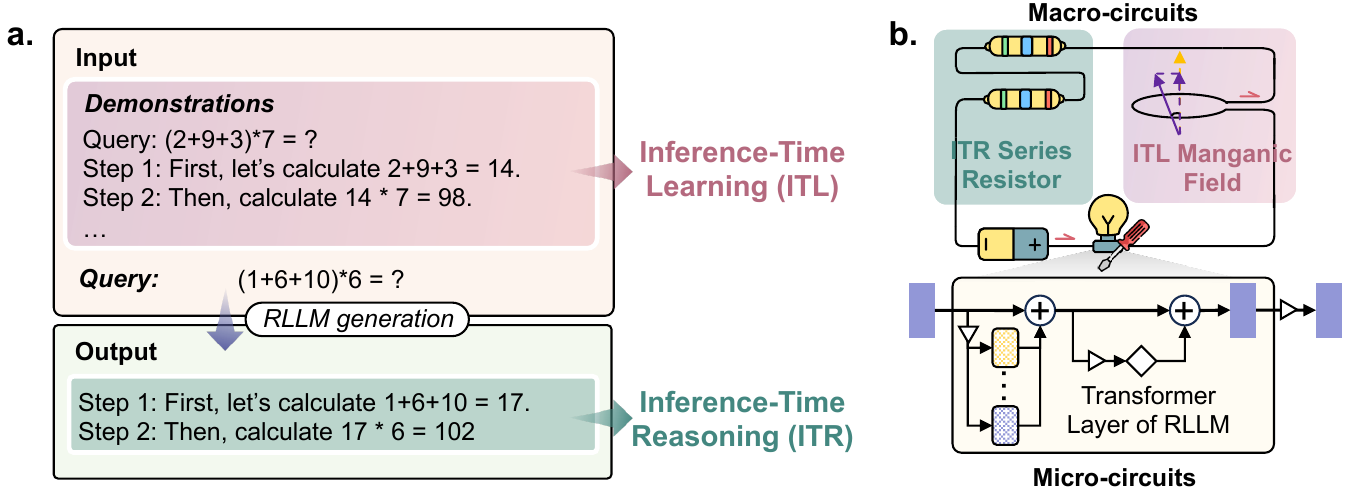}
    \caption{\textbf{The Introduction of Inference-Time Learning (ITL) and Inference-Time Reasoning (ITR), and Comparison of Micro- and Macro-Circuits.}\\
    \textit{\textbf{a,} Introduction to ITL, which pairs queries with solution in input context as references, and to ITR, which provides step-by-step reasoning in model-generated output. 
    \textbf{b,} Electronic Circuit Principles on Macro-circuit analyses: Traditional micro-circuits capture basical linguistic functions but miss higher-order reasoning principles. The micro-circuit structure follows \citet{rai2024practical}.}
    }
    \label{fig:motivation}
\end{figure*}

\vspace{-2mm}\section{Introduction}\vspace{-1mm}
Large Language Models (LLMs), like DeepSeek-R1~\citep{guo2025deepseek} and OpenAI-o1~\citep{jaech2024openai}, have recently achieved unprecedented performance in code generation, knowledge-based question answering and complex reasoning tasks~\citep{chen2025towards,li2025system,qin2024large}.
Unlike conventional software, whose behavior is explicitly encoded by human, LLMs exhibit emergent, stochastic dynamics that defy straightforward performance modeling. As a result, users often cannot anticipate when a model will succeed or fail before inference and generation, creating a ``Schrödinger’s cat'' scenario in which reliability only becomes apparent post hoc.

At inference, as shown in Figure~\ref{fig:motivation}\textit{a}, two complementary behaviors drive LLM adaptability~\citep{chen2025towards}: inference-time learning (ITL)~\citep{brown2020language,dong2022survey,qin2024what} and Inference-Time Reasoning (ITR)~\citep{nye2022show,wei2022chain,chen2024unlocking}. ITL, often termed in-context learning, enables parameter‐free generalization from examples supplied in the prompt. By contrast, ITR, frequently implemented via chain-of-thought prompting, decomposes complex tasks into interpretable reasoning steps, enhancing both accuracy and transparency. These paradigms have underpinned rapid advances in domains ranging from clinical decision support to personalized education~\citep{harrer2023attention,nazi2024large,kasneci2023chatgpt,kung2023performance}, yet they also obscure the behind rationales governing performance, impeding systematic understanding and theory-guided refinement of reasoning strategies from higher-order perspectives.

To address this gap, we present the Electronic Circuit Principle (\modelname{}), which maps ITL and ITR onto analogues of classical circuit elements. As illustrated in Fig.~\ref{fig:motivation}\textit{b}, in this principle, ITL acts as an inference-time‐varying electromotive force, enabling demonstration incorporation much like induced voltage. Meanwhile, ITR maps onto an electrical network in which each reasoning subtask contributes series resistance, such that the total resistance quantifies cumulative cognitive load. By formalizing Faraday’s and Ohm’s laws of circuits, \modelname{} furnishes a principled framework that both predicts LLM performance and guides modular optimization of reasoning components.

We validate this approach on nine widely used LLMs across 350 tasks (over 70,000 samples), achieving a Spearman correlation > 0.7 between \modelname{}‐based predictions and empirical accuracy. This represents about 60\% correlation improvement over current inference-time scaling law methods that account for post-inference results, highlighting the enhanced predictive capacity of \modelname{}.
Moreover, \modelname{}‐inspired adjustments, treating demonstration strength optimization as “voltage” improvement and reasoning strategy optimization as “resistance” reduction, consistently improve model performance based on theoretical insights.

Notably, LLMs employing \modelname{}-optimized strategies surpassed 80\% of human competitors in the International Olympiad in Informatics (IOI) and the International Mathematical Olympiad (IMO) based on much weaker LLMs. Furthermore, in subjective and exploratory domains such as academic research, \modelname{}-optimized strategies contributed to a minimum 10\% improvement in human performance metrics. These findings demonstrate the utility of the Electronic Circuit principle for targeted modular optimization.

In summary, our contributions are as follows:
\begin{itemize}[leftmargin=2ex,topsep=0pt]
    \item \textbf{Uncovering Macro-level Principles in LLMs:} We introduce the Electronic Circuit Principle (\modelname{}), a pioneering framework that redefines model performance through the lens of macro-circuit, providing a unified theoretical paradigm for quantitatively predicting and optimizing LLM performance.\vspace{-2pt}
    \item \textbf{Pioneering Accurate Performance Prediction:} By drawing analogies to Faraday's Law of Induction and Ohm's Law, we offer new insights into LLM reasoning mechanisms, enabling precise performance predictions with a Spearman correlation exceeding 70\% from 9 widely-used LLMs across more than 350 tasks.\vspace{-2pt}
    \item \textbf{Guiding Modular Performance Optimization:} Leveraging \modelname{}, we interpret existing optimization strategies and propose novel optimization strategies informed by \modelname{} theory. These strategies have led to significant breakthroughs in various fields, from mathematical and programming competitions to cutting-edge academic research.
\end{itemize}
	\begin{figure}[t]
    \centering
    \includegraphics[width=0.99\textwidth]{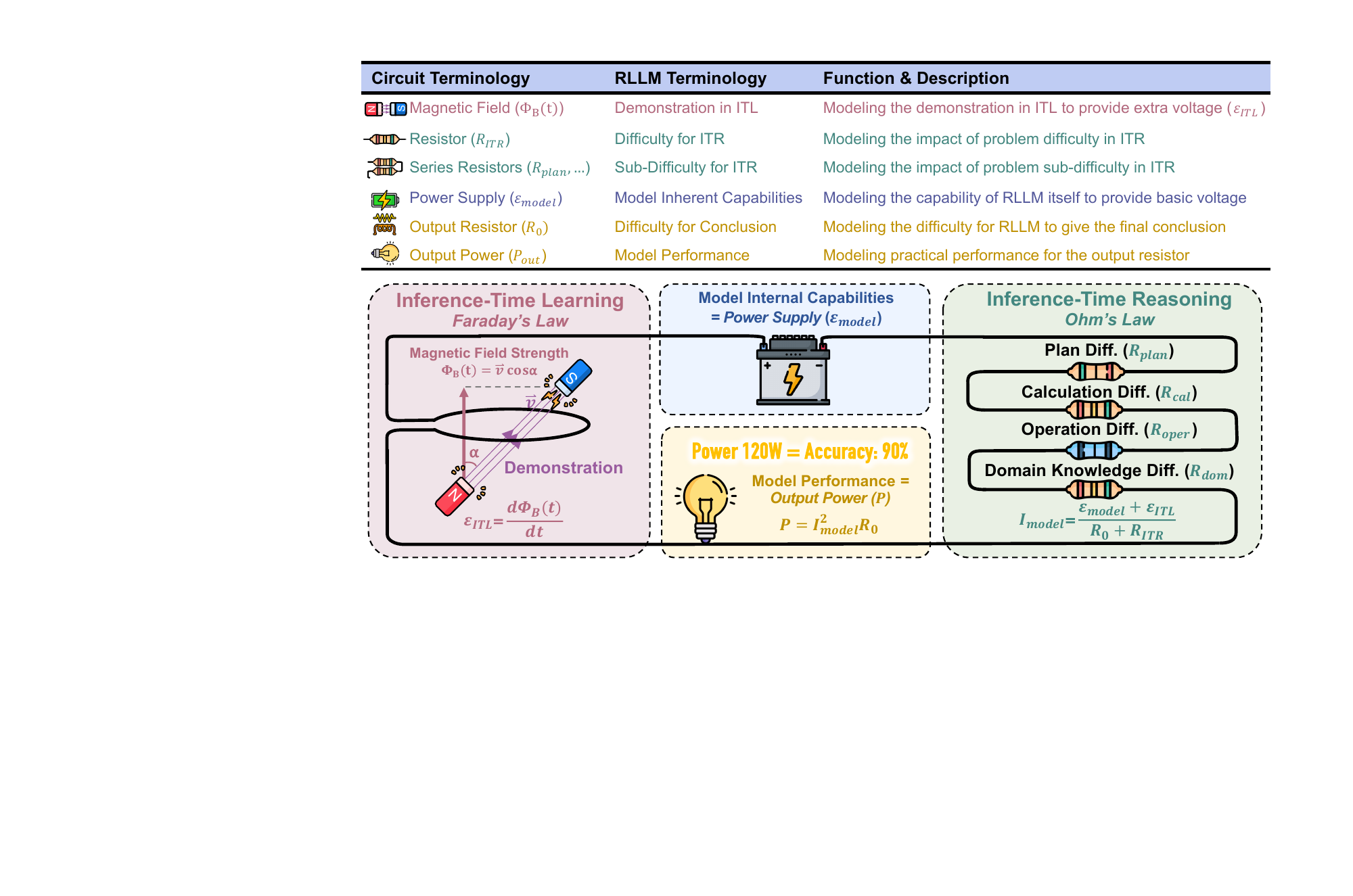}
    \caption{\textbf{Glossary of Terms and Schematic of Electronic Circuit Principles (\modelname{}).}\\
    \textit{Inference-time learning (ITL) is depicted as a dynamic semantic magnetic field whose strength rises when demonstrations accompany the prompt, adding potential that steers token selection. Inference-time reasoning (ITR) is modelled as a series circuit: each reasoning sub-task acts as a resistor, and their sum gives the total cognitive load. The LLMs' intrinsic capacity supplies the driving voltage, and empirical accuracy corresponds to the power delivered across the load. ``Strength'', ``Resistance'' and ``Voltage'' map to field intensity, sub-task difficulty and LLM capability, respectively, offering a compact framework to quantify and compare LLM behaviour.}
    }
    \label{fig:main}
\end{figure}

\vspace{-2mm}\section{Electronic Circuit Principles}\vspace{-1mm}
Electronic Circuit Principles show that model performance (\( P_{\text{out}} \)) relies not only on basic voltage from model inherent capabilities ($\mathcal{E}_{\text{model}}$) but also on external factors like reasoning difficulty ($R_{ITR}$) and additional voltage from extra demonstrations in ITL ($\mathcal{E}_{ITL}$), which are described as follows in detail:

\vspace{-2mm}\subsection{Model Inherent Capabilities as Power Supply}\vspace{-1mm}
In \modelname{}, we draw an analogy by conceptualizing LLM's inherent capabilities as a power supply that drives computational processes and reasoning tasks (see in Fig.~\ref{fig:main}). The voltage value \( \mathcal{E}_{\text{model}} \) represents the strength of the inherent capability, governing the LLM's capacity to execute a range of tasks. Just as the voltage in an electrical circuit determines the power available to drive various components, the  internal capabilities govern its ability to manage increasingly complex inputs.

\vspace{-2mm}\subsection{Faraday's Law of Inference-Time Learning}\vspace{-1mm}

Inspired by \citet{wang-etal-2023-label}, the process of ITL can be conceptualized as the interaction of LLM and contextual data, whereby the model absorbs, retains, and ultimately releases semantic information. This dynamic flow of information is analogous to the behavior of a magnetic field, where its initial strength, $\Phi^B_0$, diminishes to zero over time. Based on this analogy, as illustrated in Fig.~\ref{fig:main} (left), we introduce Faraday's Law within the context of ITL, where the rate of change in the "magnetic field" induces a corresponding voltage. Specifically, we define the voltage as:
\begin{equation}
    \mathcal{E}_{\text{ITL}} = -\frac{d\Phi^B(t)}{dt} = \lambda\Phi_0^B,
\end{equation}
where $\lambda$ denotes the uniform rate of magnetic field intensity decay. This framework provides a quantitative approach to understanding ITL by the concept of a ``semantic magnetic field''.

\vspace{-2mm}\subsection{Ohm's Law of Inference-Time Reasoning}\vspace{-1mm}
Similarly, \citet{chen2024unlocking} have derived a combination law for reasoning within LLM, which parallels the series formula for electrical resistors (See Methods).
Drawing on this, we propose a principle for the difficulty of ITR that is analogous to series circuits. In this principle, each reasoning process introduces a distinct resistor ($R_{\text{ITR}}=\sum_i R_i$), compounding the task's difficulty ($R_i$). As illustrated in Fig.~\ref{fig:main} (right), Ohm's law for ITR follows:
\begin{equation}
    I_{model} = \frac{\mathcal{E}_{\text{model}} + \mathcal{E}_{\text{ITL}}}{R_{\text{ITR}} + R_0},\label{eq:ohm}
\end{equation}
where $R_0$ represents the resistance of an output resistor with static resistance value in the circuit, which reflects the difficulty of reasoning conclusion.
This formulation quantifies the cumulative effect of reasoning resistors, providing a modular quantitative perspective on  task complexity.

\vspace{-2mm}\subsection{Model Performance as Output Power of Bulb}\vspace{-1mm}
Extending the \modelname{}, we equate model performance to the power output of a light bulb in an electronic circuit. Task execution efficiency depends on a nonlinear interaction of internal and external factors, rather than solely on the model’s intrinsic capabilities. The output power, $P_{\text{out}}$, can be formally expressed as:
\begin{equation}
    P_{\text{out}} = I^2_{model} R_0 =\frac{\left(\mathcal{E}_{\text{model}} + \mathcal{E}_{\text{ITL}}\right)^2 R_0}{(R_{\text{ITR}} + R_0)^2}.\label{eq:power}
\end{equation}
Here, output power represents the model's effective performance, with higher power indicating better task execution, akin to a brighter bulb. The equation highlights that maximizing performance requires enhancing the basic voltage of power supply ($\mathcal{E}_{\text{model}}$) of LLM, minimizing reasoning resistor ($R_{\text{ITR}}$) in ITR, and optimizing the extra voltage ($\mathcal{E}_{\text{ITL}}$) provided by ITL. This synergy explains and predicts the model’s performance on complex tasks quantitatively.
	
\begin{figure}[t]
    \centering
    \includegraphics[width=\textwidth]{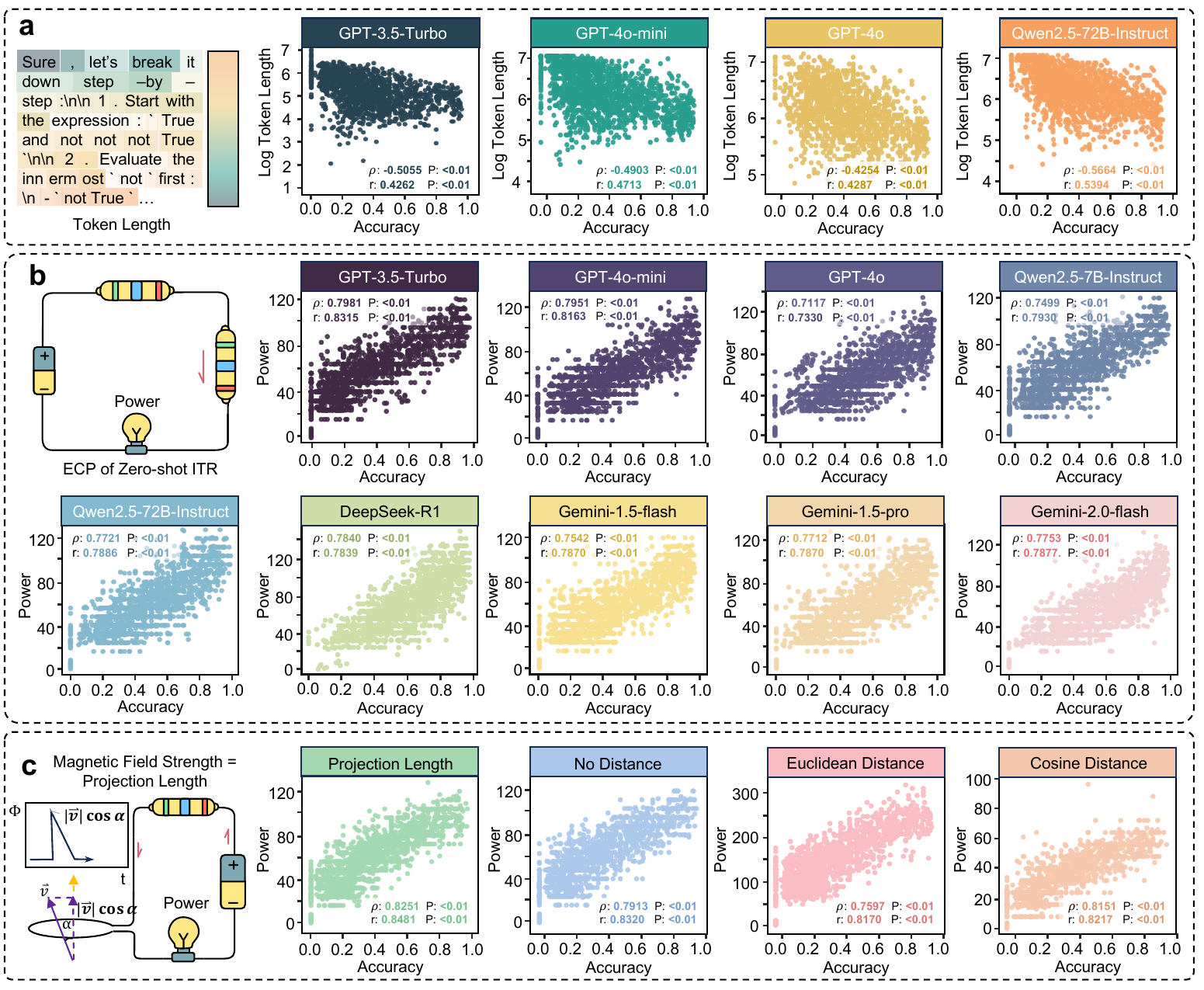}
    \caption{
      \textbf{Verification of the \modelname{} via zero‐shot reasoning.}\\
      \textit{\textbf{a,} Log-linear relationship between the classical inference-time scaling law (log token length versus task accuracy) and empirical performance of 4 LLMs, GPT-3.5 Turbo, GPT-4o-mini, GPT-4o and Qwen2.5-7B-Instruct, reveals a significant negative correlation (Spearman $\rho \in [-0.5664, -0.4254]$,$P < 0.01$).
      \textbf{b,} Observed zero-shot accuracy plotted against \modelname{}-predicted computational power for 9 LLMs shows a strong positive correlation (Spearman $\rho >0.71$, $P < 0.01$), confirming the \modelname{}'s predictive validity.
      \textbf{c,} Comparative evaluation of 4 semantic-similarity metrics used to estimate field strength: projection length, Manhattan ($L_1$) distance, Euclidean ($L_2$) distance and cosine similarity, identifies projection length as the most accurate predictor. More complete results are shown in Fig.~\ref{fig:itl-verification-append} in Appendix.}
    }

    \label{fig:verification}
    
\end{figure}

\vspace{-2mm}\section{Results}\vspace{-1mm}
\subsection{Accurate Performance Prediction}\vspace{-1mm}

\subsubsection{ITR Meets Ohm’s Law to Accurately Predict Zero-shot Performance}\vspace{-1mm}
To investigate whether ITR obeys Ohm’s law, we eliminate all inference-time learning effects  (i.e., we exclude any in-context demonstrations in the prompts), so that $\mathcal{E}_{\mathrm{ITL}} = 0$.  Since the model current \(I_{\mathrm{model}}\) in Eq.~\ref{eq:ohm} be measured directly, we instead quantify the output power to assess ITR’s adherence to Ohm’s law, which can be calculated as follows:
\begin{equation}
    Acc \propto P_{\mathrm{out}}
    = \frac{\mathcal{E}_{\mathrm{model}}^2\,R_0}{(R_{\mathrm{ITR}} + R_0)^2}.
    \label{eq:itr}
\end{equation}
As shown in Fig.~\ref{fig:verification}\textit{a}, the traditional Inference Time Scaling Law, predicting performance from token length alone, fails to predict reasoning performance accurately (Spearman’s $\rho \approx -0.5$  and Pearson’s $r < 0.53$). In contrast, as illustrated in Fig.~\ref{fig:verification}\textit{b}, the output power of our \modelname{} correlates strongly and positively with actual accuracy across every model tested ($\rho, r > 0.7$).
Our findings reveal that reasoning difficulty functions analogously to electrical resistance: as cognitive ``resistance'' increases, the ``current'' of information flow, and the relevant ``power'' of reasoning output diminishes, echoing the dynamics of Ohm’s law.

\vspace{-2mm}\subsubsection{ITL Meets Faraday’s Law to Accurately Predict Few-shot Performance}\vspace{-1mm}
Next, we introduce a novel conceptual framework for ITL, drawing an analogy to Faraday’s law: contextual demonstrations can be interpreted as generating a “semantic magnetic field” that influences task performance. To probe this analogy, we compare four proximity metrics: projection length, a baseline with no distance modelling, Euclidean distance, and cosine distance, quantifying the relationship between demonstrations and task prompts. As shown in Fig.~\ref{fig:verification}\textit{c}, projection length exhibits the strongest correlation with task accuracy (\(\rho=0.8251\), \(r=0.8481\), \(R^2=0.6095\)), outperforming both the No Distance baseline (\(\rho=0.7913\), \(r=0.8320\), \(R^2=0.5555\)) and conventional metrics such as Euclidean distance (\(\rho=0.7597\), \(r=0.8170\), \(R^2=0.6017\)) and cosine distance (\(\rho=0.8151\), \(r=0.8217\), \(R^2=0.5190\)). Strikingly, the latter two, even though widely used for demonstration retrieval, perform worse than having no distance modelling at all. These findings reveal that projection length between demonstrations and task requests is another key to effectively model ITL effectiveness.

\begin{figure*}[!b]
    \centering
    \includegraphics[width=0.99\textwidth]{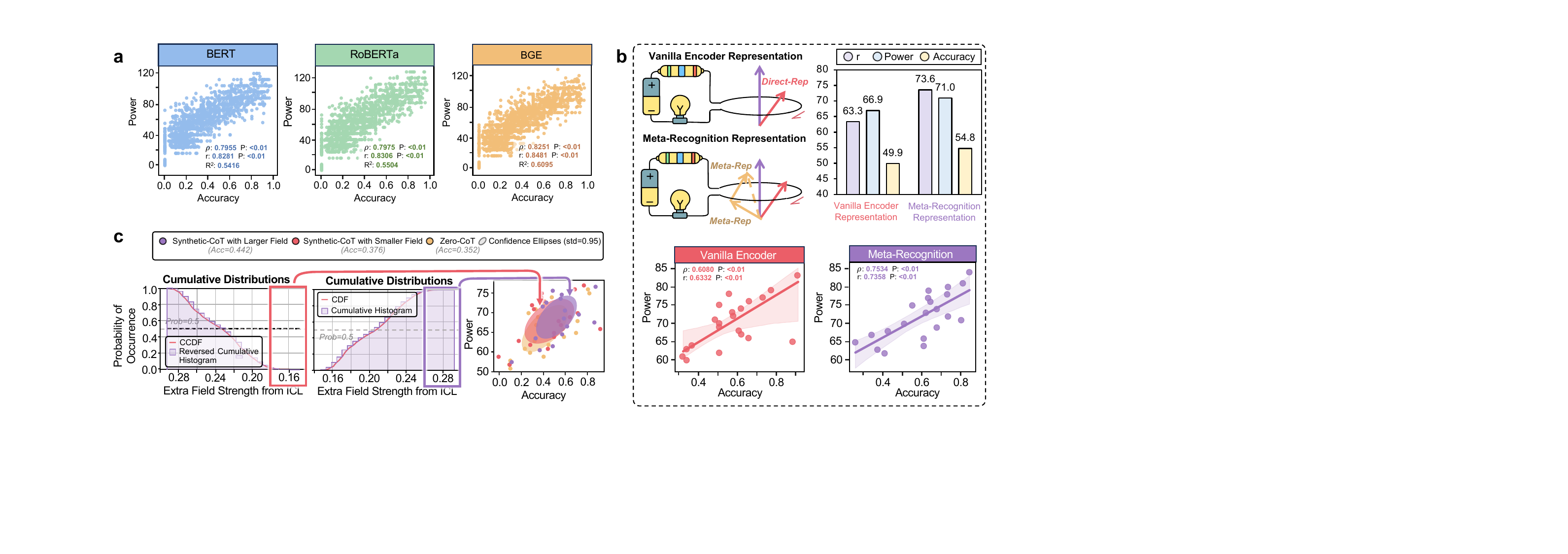}
    \caption{
      \textit{\textbf{Interpreting for Inference‐Time Learning Strategies of \modelname{}.}\\
      \textbf{a,} \modelname{}‐predicted theoretical power correlates positively with accuracy across three semantic encoders (BERT, RoBERTa, BGE) over 350 tasks.
      \textbf{b,} Meta‐recognition embeddings yield higher mean theoretical power (73.6 vs 66.9) and accuracy (54.8\% vs 49.9\%) than vanilla encoders, with stronger Spearman correlations (0.75 vs 0.61, $P<0.01$), which using RoBERTa as semantic encoders.  
      \textbf{c,} Cumulative distributions (CCDF, CDF) and 95\% confidence ellipses for three ITR configurations (Acc = 0.442, 0.376, 0.352) illustrate that larger induced semantic fields correspond to higher theoretical power and observed accuracy.}\vspace{-8pt}
    }
    \label{fig:ITL-exploration}
\end{figure*}

\vspace{-2mm}\subsection{Modular Performance Optimization}\vspace{-1mm}
\subsubsection{Interpreting Existing Optimization Strategies by Magnetic Field Adjustment}\vspace{-1mm}

\paragraph{Enhanced semantic encoders sharpen magnetic-field estimates and thereby improve retrieval accuracy.}  
Recent progress in neural semantic encoders has markedly improved the performance of ITL tasks~\citep{liu2024incontext}. To understand the underlying reason, we compared three prominent encoders (BERT~\citep{devlin-etal-2019-bert}, RoBERTa~\citep{liu2019roberta}, and the more recent BGE~\citep{chen2024bge}), by correlating power-accuracy metrics (Fig.~\ref{fig:ITL-exploration}\textit{a}).  BGE exhibited the strongest agreement with ground truth (Spearman’s $\rho = 0.83$, Pearson’s $r = 0.85$, $R^{2} = 0.61$), surpassing its predecessors and indicating that higher-quality embeddings generate more coherent semantic fields and more reliable performance forecasts.\vspace{-5pt}

\paragraph{Meta-recognition representations further augment magnetic-field modelling.}  
Leveraging GPT-4~\cite{achiam2023gpt}, Meta-Recognition~\citep{didolkar2024metacognitive} improves LLMs' performance by decomposing complex queries into modular sub-capabilities for better ITL representation.
We attribute its effectiveness to the enhanced sample representation modelling, which enhances ITL voltage and subsequently boosts output power and accuracy.
As illustrated in Fig.~\ref{fig:ITL-exploration}\textit{b}, this strategy raised theoretical power from 66.9 to 73.6 and average experimental accuracy from 49.9\% to 54.8\%. Concordance with the ground-truth distribution likewise improved ($\Delta $Spearman’s $\rho = 0.15$; $\Delta $Pearson’s $r = 0.10$; both $P < 0.01$), implying that sub-capability embeddings capture latent magnetic semantics more effectively than holistic representations.\vspace{-5pt}

\paragraph{Synthetic-CoT consistently consistently produces positive magnetic fields.}  
Synthetic-CoT requires LLMs to generate demonstrations at inference~\citep{shao2023synthetic}. Stronger fields correlate positively with retrieval accuracy, as shown by complementary cumulative distributions and 95\% confidence ellipses in power–accuracy space (Fig.~\ref{fig:ITL-exploration}\textit{c}). High-field Synthetic-CoT achieved an accuracy of 0.442, outperforming its low-field counterpart (0.376) and the zero-shot baseline (0.352). These converging results identify field magnitude as a key determinant of ITL performance and outline scalable principles for tuning magnetic fields through representation learning and prompt engineering.
\begin{figure*}[t]
    \centering
    \includegraphics[width=\textwidth]{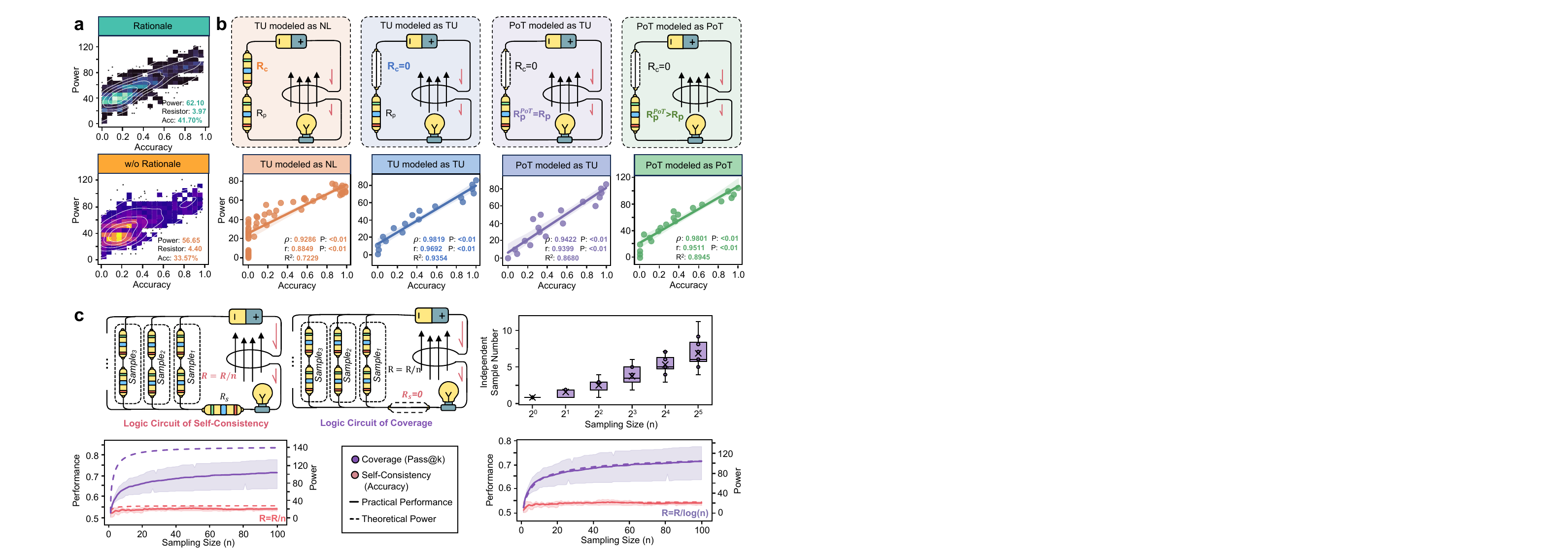}
    \caption{
      \textbf{Interpreting for Inference‐Time Learning Strategies of \modelname{}.}\\
      \textit{\textbf{a,} ITR reconfigures the model’s reasoning circuit: relative to Rationale, removing the rationale (w/o Rationale) increases effective resistance (3.97 $\Omega$ vs 4.40 $\Omega$), lowers theoretical power (62.1 vs 56.7) and improves accuracy (41.7\% vs 33.6\%), as shown by the contour shifts. \textbf{b,} Applying natural language (NL), tool usage (TU) or program-of-thought (PoT) to instantiate the ``superconducting'' phenomenon ($R_c = 0$) surpasses NL for better Power-Accuracy correlations, in line with findings on externalized reasoning. \textbf{c,} Circuit analogies for two sampling strategies. Self-consistency sampling comprises parallel reasoning paths merging through an aggregation resistance \( R_s \), denoting consensus cost; coverage sampling sets (\( R_s = 0 \)), allowing independent paths for greater diversity. \textbf{d,} Empirical trends follow a logarithmic relation with sample count. \textbf{e,} Adopting independent rather than mixture sample counts motivates replacing linear models with logarithmic ones, markedly improving predictions.}\vspace{-8pt}
}
    \label{fig:ITR-strategy}
\end{figure*}

\vspace{-2mm}\subsubsection{Interpreting Existing Optimization Strategies through Resistor Reorganization}\vspace{-1mm}

\paragraph{Direct answering increases logical resistance relative to zero-shot ITR.}  
Prompting LLMs to produce answers directly, without exposing their intermediate reasoning, imposes greater cognitive demands and reduces task performance~\citep{wei2022chain,kojima2022large}, which we attribute it to imposing additional ``logical resistance'' on the inference pathway. As illustrated in Fig.~\ref{fig:ITR-strategy}\textit{a}, we found that direct answering lowers the mean theoretical power requirement (62.10 $\rightarrow$ 56.65)but simultaneously raises effective logical resistance from 3.97 to 4.40, driven by 1.1×, 1.5× and 1.6× increases in planning, calculation and local‐operation resistances, respectively. Accuracy consequently declines from 41.7\% to 33.6\%, indicating that explicit chains of thought enhance energy efficiency and steer the model towards correct solutions.\vspace{-5pt}

\paragraph{Tool usage and program-of-thought minimise calculation and planning loads.}  
Following \citet{chen2024unlocking}, representing reasoning complexity as an ohmic load, we tested whether explicit tool calls (tool usage, TU) and step-wise code generation (program-of-thought, PoT) approach a ``superconducting'' limit. As shown in Fig.~\ref{fig:ITR-strategy}\textit{b}, setting the computational resistor to zero improved the power–accuracy fit by $>8\%$ in Pearson’s $r$ and by $>10\%$, supporting the zero‐resistance ``superconducting'' hypothesis for TU. Further reducing the planning resistor for PoT produced an even tighter correspondence ($r > 0.95$ and $R^2 > 0.89$), demonstrating PoT’s superior efficiency in converting computational throughput into correct answers.\vspace{-5pt}

\paragraph{Self-consistency and inference scaling act as parallel resistors.}  
As shown in Fig.~\ref{fig:ITR-strategy}\textit{c}, executing $n$ concurrent reasoning chains and selecting the majority answer, also known as Self-consistency~\citep{wang2023selfconsistency}, can be modelled as \(n\) parallel resistors of value \(R_{\text{ITR}}\) which reduce the effective resistance to \(R_{\text{ITR}}/n\); a verification step of resistance $R_S$ sets a lower bound (See Theoretical Proof). Inference-time scaling-law analyses~\cite{wu2024inference,openai2024o1} show that enlarging $n$ drives \(R_{\text{ITR}}/n\) towards zero, asymptotically maximising accuracy, and that a perfect verifier ($R_S=0$) outperforms self-consistency alone.

Empirically, however, Pass@$k$ diverges from theory if samples are assumed independent (Fig.~\ref{fig:ITR-strategy}\textit{c} (left)). Accounting for correlations reveals only $\log n$ independent trajectories; replacing $R_0/n$ with $R_0/\log n$ restores perfect agreement between prediction and experimental data (Fig.~\ref{fig:ITR-strategy}\textit{c} (right)), confirming that the equivalent-circuit framework captures the diminishing returns of modern inference-time scaling-laws.

\begin{figure}[t]
    \centering
    \includegraphics[width=\textwidth]{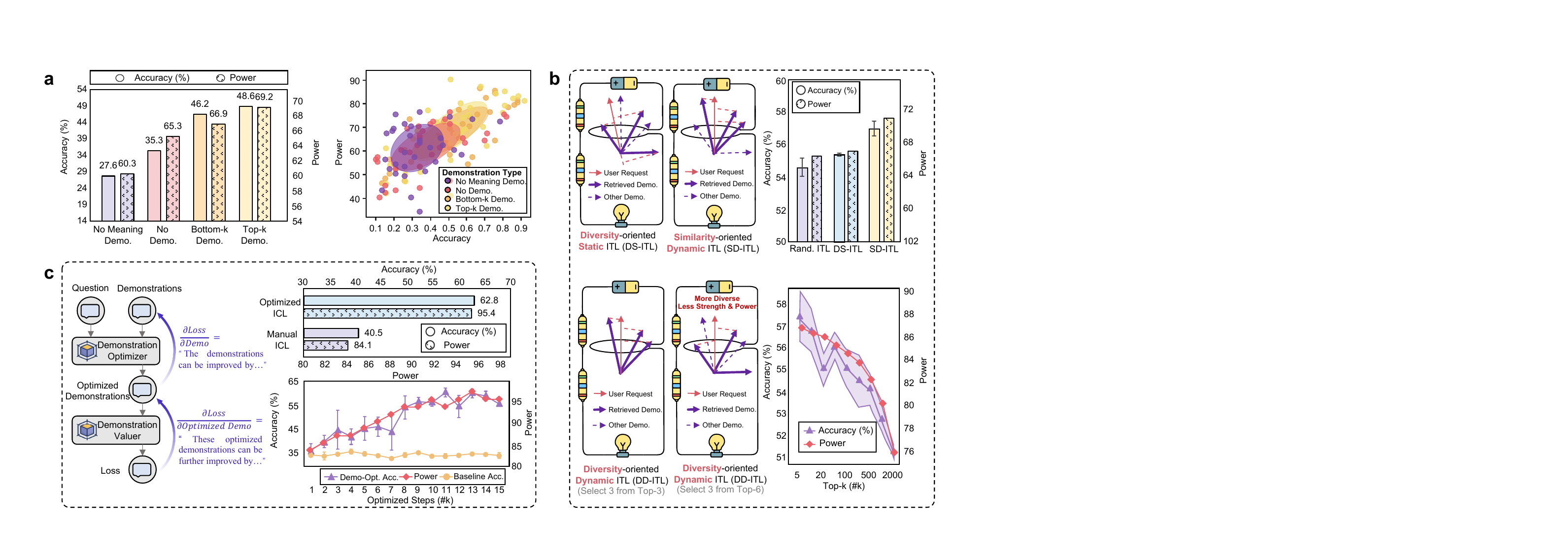}
    \caption{
      \textbf{Optimization Exploration through Magnetic Field Adjustment on \modelname{}.}\\
      \textit{\textbf{a,} Comparison of accuracy and theoretical power for no‐demo, bottom‐\(k\) and top‐\(k\) settings, top‐\(k\) retrieves for the largest extra voltage.
      \textbf{b,} Retrieval strategy comparison for random, Diversity-oriented static ICL (DS-ICL) and Similarity-oriented dynamic ICL (SD-ICL) illustrate that SD-ICL achieves the greatest accuracy and power; line plots show both metrics decline as \(k\) increases based on Diversity-oriented dynamic ICL (DD-ICL).  
        \textbf{c,}
        Compared to manual ICL, optimized demonstrations yield significantly higher accuracy and power, with consistent improvement across optimization steps.}\vspace{-8pt}
    }
    \label{fig:ITL-strategy}
\end{figure}

\vspace{-2mm}\subsection{Exploring Novel Optimization Strategies through Magnetic Field Adjustment}\vspace{-1mm}
\paragraph{When Few-shot ITR Performs Worse than Zero-shot ITR?}
LLMs exploit ITL by retrieving task demonstrations, yet the conditions under which demonstrations help remain unclear. We postulated that selecting the top-$k$ most semantically aligned examples would maximise model capacity. To test this idea we compared four retrieval policies: no demonstrations (Demo.), semantically irrelevant (no-meaning) demonstrations, bottom-$k$ and top-$k$ examples. Top-$k$ retrieval improved accuracy across tasks (Fig.~\ref{fig:ITL-strategy}\textit{a}). By contrast, no-meaning demonstrations, which exert a negative ``semantic field'', reduced effective context utilisation and yielded performance below the zero-shot baseline. These observations show that the magnetic polarity of semantic alignment determines whether demonstrations aid ICL.\vspace{-5pt}

\paragraph{The influence of diversity in ITL retrieval: a static vs. dynamic perspective.}  
Two retrieval paradigms are widely employed: static retrieval, in which a fixed demonstration set is used for all test instances, and dynamic retrieval, in which demonstrations are chosen per instance. Semantic diversity enhances static retrieval by providing complementary contextual signals~\citep{li-qiu-2023-finding},  whereas it can hinder dynamic retrieval by diluting alignment with the query~\cite{qin2024what}. We quantified these effects with random retrieval, diversity-oriented static retrieval and similarity-oriented dynamic retrieval. Similarity-oriented dynamic retrieval achieved the highest accuracy (57.0\%) and power-strength (>70) (Fig.~\ref{fig:ITL-strategy}\textit{b}). In static retrieval the same diverse set yielded higher accuracy than random retrieval, but in dynamic retrieval diversity conferred no benefit. Thus, diversity is advantageous only when demonstrations are reused across queries; when selection is instance-specific, alignment dominates.\vspace{-5pt}

\paragraph{Demo-Optimization: a novel approach to enhance demonstration efficacy in ITL.}  
Recently, gradient-based prompt optimisation can refine instructions for large language models~\citep{yuksekgonul2024textgrad}. Building on this principle we introduce ``demo-optimisation'', an algorithm that jointly adjusts the composition and ordering of demonstrations to balance their relative influence with the user prompt (See Methods in Appendix for more implementation details). As shown in Fig.~\ref{fig:ITL-strategy}\textit{c}, demo-optimisation increased accuracy by 20 percentage points relative to instruction-only optimisation and alleviated under-fitting in earlier approaches. The method therefore offers a scalable route to improve demonstration-driven reasoning.

\begin{figure}[t]
    \centering
    \includegraphics[width=\textwidth]{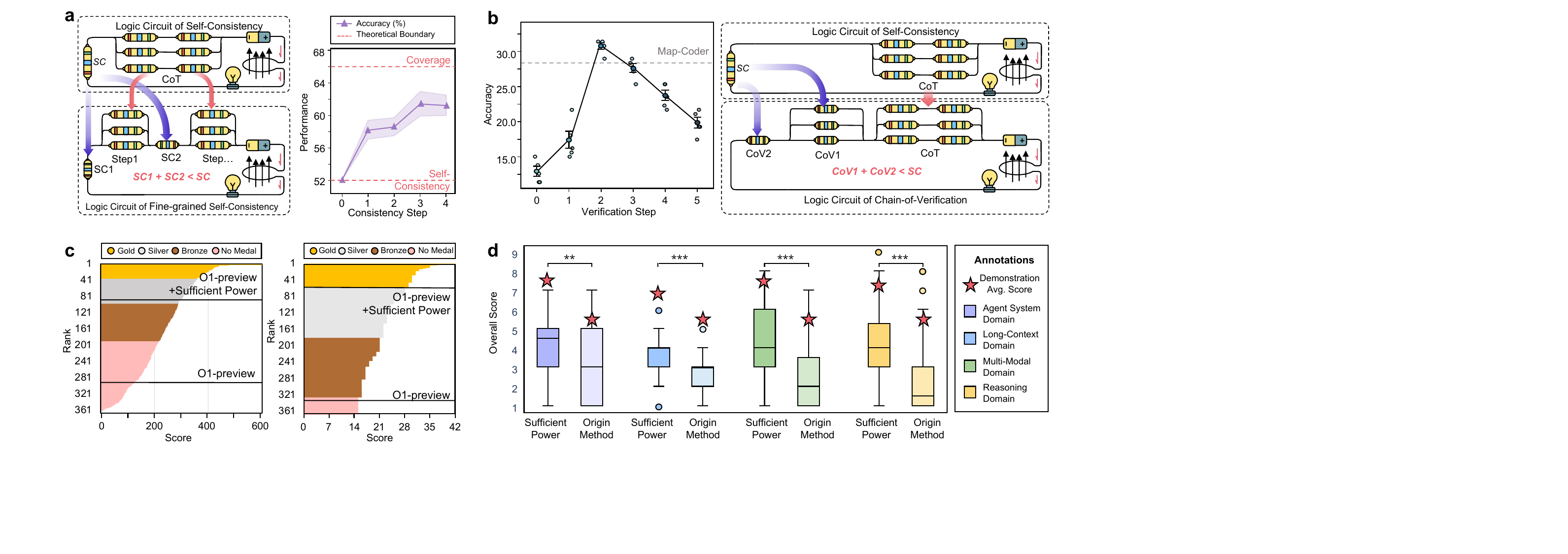}
    \caption{
        \textbf{Optimization Exploration through Resistor Reorganization(a, b) and Unified Optimization for Applications (c, d) .}\\
        \textit{\textbf{a,}
        Overview of the fine-grained self-consistency strategy in circuits. By decomposing parallel ITR resistors into smaller units, this approach effectively reduces overall circuit resistance.
        \textbf{b,}
        The Chain-of-Verification (CoV) strategy further optimizes circuit resistance by incorporating voting resistors into the self-consistency framework. This method reduces overall resistance by sampling and validating multiple voting results.
        \textbf{c,}
        Enhanced circuit performance is achieved by leveraging stronger magnetic fields and reducing logical resistance. The power-optimized model produced notable results in the IOI (left) and the IMO (right) tests.
        \textbf{d,}
        A methodology for integrating theoretical power into manual ratings during idea generation. This approach significantly improves the accuracy and reliability of the evaluation process compared to the original solution.}\vspace{-8pt}
    }
    \label{fig:ITR-optimization}
\end{figure}
\vspace{-2mm}\subsection{Exploring Novel Optimization Strategies through Resistor Reorganization}\vspace{-1mm}
\paragraph{Optimising an equal proportion of model capabilities yields a performance gain that scales with the intrinsic difficulty of each capability.}
Reducing the resistance of the most challenging reasoning component from $R$ to $\frac{R}{k}$ therefore produces a larger decrease in the total inference-time resistance, \( R_{\text{ITR}} \) , than an equivalent change applied to an easier component (See Theoretical Proof in Appendix for mathematical proof).
Meta-analysis (Methods) of benchmark problems requiring long-horizon planning confirms this prediction: the exploration-centred reasoning procedure of model o1-preview~\cite{openai2024o1} preferentially improves high-resistance components and delivers the largest gains on the most complex tasks. By contrast, tasks dominated by specialised domain knowledge benefit less, owing to limited optimisation of the corresponding knowledge component.\vspace{-5pt}

\paragraph{Fine-grained self-consistency reduces verification resistance.}
Easily, we can prove that, with a stable supply voltage, we can split a self-consistency ensemble into more fine-grained step-by-step self-consistency by conducting majority voting verification for each step, so that each resistance value of verification resistor can be smaller, and we can obtain larger output power (see Theoretical Proof). Indeed, as the number of self-consistency steps increases, the theoretical power value gradually increases. As shown in Fig.~\ref{fig:ITR-optimization}\textit{a}, the corresponding performance also gradually increases. At the same time, the corresponding upper bound and lower bound also satisfy the original self-consistency and coverage. This not only proves the effectiveness of our proposed fine-grained self-consistency method but also proves the practical usability of the theoretical model.\vspace{-5pt}

\paragraph{Chain-of-verification balances parallel and serial checks.}
Traditional self-consistency frameworks rely on a sequential verification, where each step depends on the completion of the previous one. This structure often incurs high computational costs, as subsequent tasks cannot begin until earlier ones are resolved. To overcome this limitation, we introduce the ``chain-of-verification (CoV)," a novel strategy that parallelizes verification tasks into independent components. Each component operates concurrently, validating distinct parts of the system without interdependence. This parallel approach reduces overall computational resistance by distributing the workload across multiple verifiers, thereby enabling more effective validation. As illustrated in Fig.~\ref{fig:ITR-optimization}\textit{b}, parallelization not only accelerates the verification process but also enhances model performance by mitigating bottlenecks. This improvement is particularly pronounced in large-scale systems, where sequential verification would otherwise impede efficiency. The effectiveness of CoV can also be mathematically proved in Theoretical Proof.

\vspace{-2mm}\subsection{Unified Optimization for Downstream Applications}\vspace{-1mm}

Adequate computational power is a primary determinant of LLM performance across disparate tasks. We benchmarked an enhanced version of the o1-preview model on algorithmic problem solving (International Olympiad in Informatics, IOI), mathematical reasoning (International Mathematical Olympiad, IMO) and open-ended idea generation in the AI-Scientist framework~\cite{lu2024aiscientist}, using the ensemble procedure described above.

Increasing the output power yielded marked gains (Figure \ref{fig:ITR-optimization}\textit{c}): on the IOI code\_contests~\citep{doi:10.1126/science.abq1158} test set the power-augmented model outperformed AlphaCode~\citep{doi:10.1126/science.abq1158} and all other published baselines. In live competitions (IOI 2024 and IMO 2024) it surpassed \~80\% of human participants, earning silver and bronze medals, respectively, and highlighting compute as a critical factor when conventional optimisation is insufficient.

Enhanced compute also improved performance in creativity-oriented tasks. In the AI-Scientist benchmark, the same configuration increased the mean expert-rated novelty and impact scores by >1.0 (Fig.~\ref{fig:ITR-optimization}\textit{d}), indicating that scaling compute extends the applicability of LLMs beyond correctness-centred objectives.

\vspace{-2mm}\section{Conclusion \& Discussion}\vspace{-1mm}
The Electronic Circuit Principles (\modelname{}) introduces a unified principle for understanding and optimizing LLMs' performance. By quantifying internal dynamics, \modelname{} clarifies model behavior and guides modular performance improvements. Our experiments confirm that \modelname{} reliably captures model complexities, providing a robust tool for AI analysis and enhancement. Moreover, \modelname{}-driven power optimizations applied to the International Olympiad in Informatics and the International Mathematical Olympiad outperform current leading methods. These findings validate \modelname{}’s theoretical foundation and underscore its potential to elevate AI reasoning and learning efficiency in next-generation systems.
\vspace{-2mm}\section{Code Availability}\vspace{-1mm}
The code and relevant data are available at \url{https://github.com/LightChen233/ECP}.

	\newpage
	\bibliographystyle{plainnat}
	\bibliography{custom}

\begin{thebibliography}{46}
\providecommand{\natexlab}[1]{#1}
\providecommand{\url}[1]{\texttt{#1}}
\expandafter\ifx\csname urlstyle\endcsname\relax
  \providecommand{\doi}[1]{doi: #1}\else
  \providecommand{\doi}{doi: \begingroup \urlstyle{rm}\Url}\fi

\bibitem[Achiam et~al.(2023)Achiam, Adler, Agarwal, Ahmad, Akkaya, Aleman, Almeida, Altenschmidt, Altman, Anadkat, et~al.]{achiam2023gpt}
Josh Achiam, Steven Adler, Sandhini Agarwal, Lama Ahmad, Ilge Akkaya, Florencia~Leoni Aleman, Diogo Almeida, Janko Altenschmidt, Sam Altman, Shyamal Anadkat, et~al.
\newblock Gpt-4 technical report.
\newblock \emph{arXiv preprint arXiv:2303.08774}, 2023.

\bibitem[Brown et~al.(2020)Brown, Mann, Ryder, Subbiah, Kaplan, Dhariwal, Neelakantan, Shyam, Sastry, Askell, et~al.]{brown2020language}
Tom Brown, Benjamin Mann, Nick Ryder, Melanie Subbiah, Jared~D Kaplan, Prafulla Dhariwal, Arvind Neelakantan, Pranav Shyam, Girish Sastry, Amanda Askell, et~al.
\newblock Language models are few-shot learners.
\newblock \emph{Advances in neural information processing systems}, 33:\penalty0 1877--1901, 2020.

\bibitem[Chen et~al.(2024{\natexlab{a}})Chen, Xiao, Zhang, Luo, Lian, and Liu]{chen2024bge}
Jianlv Chen, Shitao Xiao, Peitian Zhang, Kun Luo, Defu Lian, and Zheng Liu.
\newblock Bge m3-embedding: Multi-lingual, multi-functionality, multi-granularity text embeddings through self-knowledge distillation.
\newblock \emph{arXiv preprint arXiv:2402.03216}, 2024{\natexlab{a}}.

\bibitem[Chen et~al.(2024{\natexlab{b}})Chen, Qin, WANG, Zhou, and Che]{chen2024unlocking}
Qiguang Chen, Libo Qin, Jiaqi WANG, Jingxuan Zhou, and Wanxiang Che.
\newblock Unlocking the capabilities of thought: A reasoning boundary framework to quantify and optimize chain-of-thought.
\newblock In \emph{The Thirty-eighth Annual Conference on Neural Information Processing Systems}, 2024{\natexlab{b}}.
\newblock URL \url{https://openreview.net/forum?id=pC44UMwy2v}.

\bibitem[Chen et~al.(2025{\natexlab{a}})Chen, Qin, Liu, Liao, Wang, Zhou, and Che]{chen2025rbf++}
Qiguang Chen, Libo Qin, Jinhao Liu, Yue Liao, Jiaqi Wang, Jingxuan Zhou, and Wanxiang Che.
\newblock Rbf++: Quantifying and optimizing reasoning boundaries across measurable and unmeasurable capabilities for chain-of-thought reasoning.
\newblock \emph{arXiv preprint arXiv:2505.13307}, 2025{\natexlab{a}}.

\bibitem[Chen et~al.(2025{\natexlab{b}})Chen, Qin, Liu, Peng, Guan, Wang, Hu, Zhou, Gao, and Che]{chen2025towards}
Qiguang Chen, Libo Qin, Jinhao Liu, Dengyun Peng, Jiannan Guan, Peng Wang, Mengkang Hu, Yuhang Zhou, Te~Gao, and Wangxiang Che.
\newblock Towards reasoning era: A survey of long chain-of-thought for reasoning large language models.
\newblock \emph{arXiv preprint arXiv:2503.09567}, 2025{\natexlab{b}}.

\bibitem[Cheng et~al.(2024)Cheng, Huang, and Wei]{cheng2024adapting}
Daixuan Cheng, Shaohan Huang, and Furu Wei.
\newblock Adapting large language models via reading comprehension.
\newblock In \emph{The Twelfth International Conference on Learning Representations}, 2024.
\newblock URL \url{https://openreview.net/forum?id=y886UXPEZ0}.

\bibitem[Cobbe et~al.(2021)Cobbe, Kosaraju, Bavarian, Chen, Jun, Kaiser, Plappert, Tworek, Hilton, Nakano, et~al.]{cobbe2021training}
Karl Cobbe, Vineet Kosaraju, Mohammad Bavarian, Mark Chen, Heewoo Jun, Lukasz Kaiser, Matthias Plappert, Jerry Tworek, Jacob Hilton, Reiichiro Nakano, et~al.
\newblock Training verifiers to solve math word problems.
\newblock \emph{arXiv preprint arXiv:2110.14168}, 2021.

\bibitem[Devlin et~al.(2019)Devlin, Chang, Lee, and Toutanova]{devlin-etal-2019-bert}
Jacob Devlin, Ming-Wei Chang, Kenton Lee, and Kristina Toutanova.
\newblock {BERT}: Pre-training of deep bidirectional transformers for language understanding.
\newblock In \emph{Proceedings of the 2019 Conference of the North {A}merican Chapter of the Association for Computational Linguistics: Human Language Technologies, Volume 1 (Long and Short Papers)}, pages 4171--4186, Minneapolis, Minnesota, June 2019. Association for Computational Linguistics.
\newblock \doi{10.18653/v1/N19-1423}.
\newblock URL \url{https://aclanthology.org/N19-1423}.

\bibitem[Didolkar et~al.(2024)Didolkar, Goyal, Ke, Guo, Valko, Lillicrap, Rezende, Bengio, Mozer, and Arora]{didolkar2024metacognitive}
Aniket~Rajiv Didolkar, Anirudh Goyal, Nan~Rosemary Ke, Siyuan Guo, Michal Valko, Timothy~P Lillicrap, Danilo~Jimenez Rezende, Yoshua Bengio, Michael~Curtis Mozer, and Sanjeev Arora.
\newblock Metacognitive capabilities of {LLM}s: An exploration in mathematical problem solving.
\newblock In \emph{AI for Math Workshop @ ICML 2024}, 2024.
\newblock URL \url{https://openreview.net/forum?id=0MsI3bSmmD}.

\bibitem[Dong et~al.(2022)Dong, Li, Dai, Zheng, Wu, Chang, Sun, Xu, and Sui]{dong2022survey}
Qingxiu Dong, Lei Li, Damai Dai, Ce~Zheng, Zhiyong Wu, Baobao Chang, Xu~Sun, Jingjing Xu, and Zhifang Sui.
\newblock A survey on in-context learning.
\newblock \emph{arXiv preprint arXiv:2301.00234}, 2022.

\bibitem[Du et~al.(2025)Du, Yao, Ma, Wang, Zheng, Zhu, Liu, Liang, Jin, Wei, et~al.]{du2025supergpqa}
Xinrun Du, Yifan Yao, Kaijing Ma, Bingli Wang, Tianyu Zheng, King Zhu, Minghao Liu, Yiming Liang, Xiaolong Jin, Zhenlin Wei, et~al.
\newblock Supergpqa: Scaling llm evaluation across 285 graduate disciplines.
\newblock \emph{arXiv preprint arXiv:2502.14739}, 2025.

\bibitem[Dubey et~al.(2024)Dubey, Jauhri, Pandey, Kadian, Al-Dahle, Letman, Mathur, Schelten, Yang, Fan, et~al.]{dubey2024llama}
Abhimanyu Dubey, Abhinav Jauhri, Abhinav Pandey, Abhishek Kadian, Ahmad Al-Dahle, Aiesha Letman, Akhil Mathur, Alan Schelten, Amy Yang, Angela Fan, et~al.
\newblock The llama 3 herd of models.
\newblock \emph{arXiv preprint arXiv:2407.21783}, 2024.

\bibitem[Guo et~al.(2025)Guo, Yang, Zhang, Song, Zhang, Xu, Zhu, Ma, Wang, Bi, et~al.]{guo2025deepseek}
Daya Guo, Dejian Yang, Haowei Zhang, Junxiao Song, Ruoyu Zhang, Runxin Xu, Qihao Zhu, Shirong Ma, Peiyi Wang, Xiao Bi, et~al.
\newblock Deepseek-r1: Incentivizing reasoning capability in llms via reinforcement learning.
\newblock \emph{arXiv preprint arXiv:2501.12948}, 2025.

\bibitem[Harrer(2023)]{harrer2023attention}
Stefan Harrer.
\newblock Attention is not all you need: the complicated case of ethically using large language models in healthcare and medicine.
\newblock \emph{EBioMedicine}, 90, 2023.

\bibitem[Hendrycks et~al.(2021)Hendrycks, Burns, Basart, Zou, Mazeika, Song, and Steinhardt]{hendrycks2021measuring}
Dan Hendrycks, Collin Burns, Steven Basart, Andy Zou, Mantas Mazeika, Dawn Song, and Jacob Steinhardt.
\newblock Measuring massive multitask language understanding.
\newblock In \emph{International Conference on Learning Representations}, 2021.
\newblock URL \url{https://openreview.net/forum?id=d7KBjmI3GmQ}.

\bibitem[Jaech et~al.(2024)Jaech, Kalai, Lerer, Richardson, El-Kishky, Low, Helyar, Madry, Beutel, Carney, et~al.]{jaech2024openai}
Aaron Jaech, Adam Kalai, Adam Lerer, Adam Richardson, Ahmed El-Kishky, Aiden Low, Alec Helyar, Aleksander Madry, Alex Beutel, Alex Carney, et~al.
\newblock Openai o1 system card.
\newblock \emph{arXiv preprint arXiv:2412.16720}, 2024.

\bibitem[Jin et~al.(2021)Jin, Pan, Oufattole, Weng, Fang, and Szolovits]{jin2021disease}
Di~Jin, Eileen Pan, Nassim Oufattole, Wei-Hung Weng, Hanyi Fang, and Peter Szolovits.
\newblock What disease does this patient have? a large-scale open domain question answering dataset from medical exams.
\newblock \emph{Applied Sciences}, 11\penalty0 (14):\penalty0 6421, 2021.

\bibitem[Jin et~al.(2019)Jin, Dhingra, Liu, Cohen, and Lu]{jin-etal-2019-pubmedqa}
Qiao Jin, Bhuwan Dhingra, Zhengping Liu, William Cohen, and Xinghua Lu.
\newblock {P}ub{M}ed{QA}: A dataset for biomedical research question answering.
\newblock In Kentaro Inui, Jing Jiang, Vincent Ng, and Xiaojun Wan, editors, \emph{Proceedings of the 2019 Conference on Empirical Methods in Natural Language Processing and the 9th International Joint Conference on Natural Language Processing (EMNLP-IJCNLP)}, pages 2567--2577, Hong Kong, China, November 2019. Association for Computational Linguistics.
\newblock \doi{10.18653/v1/D19-1259}.
\newblock URL \url{https://aclanthology.org/D19-1259/}.

\bibitem[Kasneci et~al.(2023)Kasneci, Se{\ss}ler, K{\"u}chemann, Bannert, Dementieva, Fischer, Gasser, Groh, G{\"u}nnemann, H{\"u}llermeier, et~al.]{kasneci2023chatgpt}
Enkelejda Kasneci, Kathrin Se{\ss}ler, Stefan K{\"u}chemann, Maria Bannert, Daryna Dementieva, Frank Fischer, Urs Gasser, Georg Groh, Stephan G{\"u}nnemann, Eyke H{\"u}llermeier, et~al.
\newblock Chatgpt for good? on opportunities and challenges of large language models for education.
\newblock \emph{Learning and individual differences}, 103:\penalty0 102274, 2023.

\bibitem[Kojima et~al.(2022)Kojima, Gu, Reid, Matsuo, and Iwasawa]{kojima2022large}
Takeshi Kojima, Shixiang~Shane Gu, Machel Reid, Yutaka Matsuo, and Yusuke Iwasawa.
\newblock Large language models are zero-shot reasoners.
\newblock In Alice~H. Oh, Alekh Agarwal, Danielle Belgrave, and Kyunghyun Cho, editors, \emph{Advances in Neural Information Processing Systems}, 2022.
\newblock URL \url{https://openreview.net/forum?id=e2TBb5y0yFf}.

\bibitem[Kung et~al.(2023)Kung, Cheatham, Medenilla, Sillos, De~Leon, Elepa{\~n}o, Madriaga, Aggabao, Diaz-Candido, Maningo, et~al.]{kung2023performance}
Tiffany~H Kung, Morgan Cheatham, Arielle Medenilla, Czarina Sillos, Lorie De~Leon, Camille Elepa{\~n}o, Maria Madriaga, Rimel Aggabao, Giezel Diaz-Candido, James Maningo, et~al.
\newblock Performance of chatgpt on usmle: potential for ai-assisted medical education using large language models.
\newblock \emph{PLoS digital health}, 2\penalty0 (2):\penalty0 e0000198, 2023.

\bibitem[Li and Qiu(2023)]{li-qiu-2023-finding}
Xiaonan Li and Xipeng Qiu.
\newblock Finding support examples for in-context learning.
\newblock In Houda Bouamor, Juan Pino, and Kalika Bali, editors, \emph{Findings of the Association for Computational Linguistics: EMNLP 2023}, pages 6219--6235, Singapore, December 2023. Association for Computational Linguistics.
\newblock \doi{10.18653/v1/2023.findings-emnlp.411}.
\newblock URL \url{https://aclanthology.org/2023.findings-emnlp.411}.

\bibitem[Li et~al.(2022)Li, Choi, Chung, Kushman, Schrittwieser, Leblond, Eccles, Keeling, Gimeno, Lago, Hubert, Choy, de~Masson~d’Autume, Babuschkin, Chen, Huang, Welbl, Gowal, Cherepanov, Molloy, Mankowitz, Robson, Kohli, de~Freitas, Kavukcuoglu, and Vinyals]{doi:10.1126/science.abq1158}
Yujia Li, David Choi, Junyoung Chung, Nate Kushman, Julian Schrittwieser, Rémi Leblond, Tom Eccles, James Keeling, Felix Gimeno, Agustin~Dal Lago, Thomas Hubert, Peter Choy, Cyprien de~Masson~d’Autume, Igor Babuschkin, Xinyun Chen, Po-Sen Huang, Johannes Welbl, Sven Gowal, Alexey Cherepanov, James Molloy, Daniel~J. Mankowitz, Esme~Sutherland Robson, Pushmeet Kohli, Nando de~Freitas, Koray Kavukcuoglu, and Oriol Vinyals.
\newblock Competition-level code generation with alphacode.
\newblock \emph{Science}, 378\penalty0 (6624):\penalty0 1092--1097, 2022.
\newblock \doi{10.1126/science.abq1158}.
\newblock URL \url{https://www.science.org/doi/abs/10.1126/science.abq1158}.

\bibitem[Li et~al.(2025)Li, Zhang, Zhang, Zhang, Liu, Yao, Xu, Zheng, Wang, Chen, et~al.]{li2025system}
Zhong-Zhi Li, Duzhen Zhang, Ming-Liang Zhang, Jiaxin Zhang, Zengyan Liu, Yuxuan Yao, Haotian Xu, Junhao Zheng, Pei-Jie Wang, Xiuyi Chen, et~al.
\newblock From system 1 to system 2: A survey of reasoning large language models.
\newblock \emph{arXiv preprint arXiv:2502.17419}, 2025.

\bibitem[Liu et~al.(2024)Liu, Ye, Xing, and Zou]{liu2024incontext}
Sheng Liu, Haotian Ye, Lei Xing, and James~Y. Zou.
\newblock In-context vectors: Making in context learning more effective and controllable through latent space steering.
\newblock In \emph{Forty-first International Conference on Machine Learning}, 2024.
\newblock URL \url{https://openreview.net/forum?id=dJTChKgv3a}.

\bibitem[Liu et~al.(2019)Liu, Ott, Goyal, Du, Joshi, Chen, Levy, Lewis, Zettlemoyer, and Stoyanov]{liu2019roberta}
Yinhan Liu, Myle Ott, Naman Goyal, Jingfei Du, Mandar Joshi, Danqi Chen, Omer Levy, Mike Lewis, Luke Zettlemoyer, and Veselin Stoyanov.
\newblock Roberta: A robustly optimized bert pretraining approach, 2019.
\newblock URL \url{https://arxiv.org/abs/1907.11692}.

\bibitem[Lu et~al.(2024)Lu, Lu, Lange, Foerster, Clune, and Ha]{lu2024aiscientist}
Chris Lu, Cong Lu, Robert~Tjarko Lange, Jakob Foerster, Jeff Clune, and David Ha.
\newblock The {AI} {S}cientist: Towards fully automated open-ended scientific discovery.
\newblock \emph{arXiv preprint arXiv:2408.06292}, 2024.

\bibitem[Nazi and Peng(2024)]{nazi2024large}
Zabir~Al Nazi and Wei Peng.
\newblock Large language models in healthcare and medical domain: A review.
\newblock In \emph{Informatics}, volume~11, page~57. MDPI, 2024.

\bibitem[Nye et~al.(2022)Nye, Andreassen, Gur-Ari, Michalewski, Austin, Bieber, Dohan, Lewkowycz, Bosma, Luan, et~al.]{nye2022show}
Maxwell Nye, Anders~Johan Andreassen, Guy Gur-Ari, Henryk Michalewski, Jacob Austin, David Bieber, David Dohan, Aitor Lewkowycz, Maarten Bosma, David Luan, et~al.
\newblock Show your work: Scratchpads for intermediate computation with language models.
\newblock In \emph{Deep Learning for Code Workshop}, 2022.

\bibitem[OpenAI(2024)]{openai2024o1}
OpenAI.
\newblock Learning to reason with llms.
\newblock 2024.
\newblock URL \url{https://openai.com/index/learning-to-reason-with-llms/}.

\bibitem[Pal et~al.(2022)Pal, Umapathi, and Sankarasubbu]{pal2022medmcqa}
Ankit Pal, Logesh~Kumar Umapathi, and Malaikannan Sankarasubbu.
\newblock Medmcqa: A large-scale multi-subject multi-choice dataset for medical domain question answering.
\newblock In Gerardo Flores, George~H Chen, Tom Pollard, Joyce~C Ho, and Tristan Naumann, editors, \emph{Proceedings of the Conference on Health, Inference, and Learning}, volume 174 of \emph{Proceedings of Machine Learning Research}, pages 248--260. PMLR, 07--08 Apr 2022.
\newblock URL \url{https://proceedings.mlr.press/v174/pal22a.html}.

\bibitem[Qin et~al.(2024{\natexlab{a}})Qin, Chen, Fei, Chen, Li, and Che]{qin2024what}
Libo Qin, Qiguang Chen, Hao Fei, Zhi Chen, Min Li, and Wanxiang Che.
\newblock What factors affect multi-modal in-context learning? an in-depth exploration.
\newblock In \emph{The Thirty-eighth Annual Conference on Neural Information Processing Systems}, 2024{\natexlab{a}}.
\newblock URL \url{https://openreview.net/forum?id=REVdYKGcfb}.

\bibitem[Qin et~al.(2024{\natexlab{b}})Qin, Chen, Feng, Wu, Zhang, Li, Li, Che, and Yu]{qin2024large}
Libo Qin, Qiguang Chen, Xiachong Feng, Yang Wu, Yongheng Zhang, Yinghui Li, Min Li, Wanxiang Che, and Philip~S. Yu.
\newblock Large language models meet nlp: A survey, 2024{\natexlab{b}}.

\bibitem[Rai et~al.(2024)Rai, Zhou, Feng, Saparov, and Yao]{rai2024practical}
Daking Rai, Yilun Zhou, Shi Feng, Abulhair Saparov, and Ziyu Yao.
\newblock A practical review of mechanistic interpretability for transformer-based language models.
\newblock \emph{arXiv preprint arXiv:2407.02646}, 2024.

\bibitem[Rein et~al.(2024)Rein, Hou, Stickland, Petty, Pang, Dirani, Michael, and Bowman]{rein2024gpqa}
David Rein, Betty~Li Hou, Asa~Cooper Stickland, Jackson Petty, Richard~Yuanzhe Pang, Julien Dirani, Julian Michael, and Samuel~R. Bowman.
\newblock {GPQA}: A graduate-level google-proof q\&a benchmark.
\newblock In \emph{First Conference on Language Modeling}, 2024.
\newblock URL \url{https://openreview.net/forum?id=Ti67584b98}.

\bibitem[Shao et~al.(2023)Shao, Gong, Shen, Huang, Duan, and Chen]{shao2023synthetic}
Zhihong Shao, Yeyun Gong, Yelong Shen, Minlie Huang, Nan Duan, and Weizhu Chen.
\newblock Synthetic prompting: Generating chain-of-thought demonstrations for large language models.
\newblock In \emph{International Conference on Machine Learning}, pages 30706--30775. PMLR, 2023.

\bibitem[Suzgun et~al.(2023)Suzgun, Scales, Sch{\"a}rli, Gehrmann, Tay, Chung, Chowdhery, Le, Chi, Zhou, and Wei]{suzgun-etal-2023-challenging}
Mirac Suzgun, Nathan Scales, Nathanael Sch{\"a}rli, Sebastian Gehrmann, Yi~Tay, Hyung~Won Chung, Aakanksha Chowdhery, Quoc Le, Ed~Chi, Denny Zhou, and Jason Wei.
\newblock Challenging {BIG}-bench tasks and whether chain-of-thought can solve them.
\newblock In Anna Rogers, Jordan Boyd-Graber, and Naoaki Okazaki, editors, \emph{Findings of the Association for Computational Linguistics: ACL 2023}, pages 13003--13051, Toronto, Canada, July 2023. Association for Computational Linguistics.
\newblock \doi{10.18653/v1/2023.findings-acl.824}.
\newblock URL \url{https://aclanthology.org/2023.findings-acl.824/}.

\bibitem[Valmeekam et~al.(2024)Valmeekam, Stechly, and Kambhampati]{valmeekam2024llms}
Karthik Valmeekam, Kaya Stechly, and Subbarao Kambhampati.
\newblock {LLM}s still can't plan; can {LRM}s? a preliminary evaluation of open{AI}'s o1 on planbench.
\newblock In \emph{NeurIPS 2024 Workshop on Open-World Agents}, 2024.
\newblock URL \url{https://openreview.net/forum?id=Gcr1Lx4Koz}.

\bibitem[Wang et~al.(2023{\natexlab{a}})Wang, Li, Dai, Chen, Zhou, Meng, Zhou, and Sun]{wang-etal-2023-label}
Lean Wang, Lei Li, Damai Dai, Deli Chen, Hao Zhou, Fandong Meng, Jie Zhou, and Xu~Sun.
\newblock Label words are anchors: An information flow perspective for understanding in-context learning.
\newblock In Houda Bouamor, Juan Pino, and Kalika Bali, editors, \emph{Proceedings of the 2023 Conference on Empirical Methods in Natural Language Processing}, pages 9840--9855, Singapore, December 2023{\natexlab{a}}. Association for Computational Linguistics.
\newblock \doi{10.18653/v1/2023.emnlp-main.609}.
\newblock URL \url{https://aclanthology.org/2023.emnlp-main.609}.

\bibitem[Wang et~al.(2023{\natexlab{b}})Wang, Wei, Schuurmans, Le, Chi, Narang, Chowdhery, and Zhou]{wang2023selfconsistency}
Xuezhi Wang, Jason Wei, Dale Schuurmans, Quoc~V Le, Ed~H. Chi, Sharan Narang, Aakanksha Chowdhery, and Denny Zhou.
\newblock Self-consistency improves chain of thought reasoning in language models.
\newblock In \emph{The Eleventh International Conference on Learning Representations}, 2023{\natexlab{b}}.
\newblock URL \url{https://openreview.net/forum?id=1PL1NIMMrw}.

\bibitem[Wang et~al.(2024)Wang, Ma, Zhang, Ni, Chandra, Guo, Ren, Arulraj, He, Jiang, Li, Ku, Wang, Zhuang, Fan, Yue, and Chen]{wang2024mmlupro}
Yubo Wang, Xueguang Ma, Ge~Zhang, Yuansheng Ni, Abhranil Chandra, Shiguang Guo, Weiming Ren, Aaran Arulraj, Xuan He, Ziyan Jiang, Tianle Li, Max Ku, Kai Wang, Alex Zhuang, Rongqi Fan, Xiang Yue, and Wenhu Chen.
\newblock {MMLU}-pro: A more robust and challenging multi-task language understanding benchmark.
\newblock In \emph{The Thirty-eight Conference on Neural Information Processing Systems Datasets and Benchmarks Track}, 2024.
\newblock URL \url{https://openreview.net/forum?id=y10DM6R2r3}.

\bibitem[Wei et~al.(2022)Wei, Wang, Schuurmans, Bosma, brian ichter, Xia, Chi, Le, and Zhou]{wei2022chain}
Jason Wei, Xuezhi Wang, Dale Schuurmans, Maarten Bosma, brian ichter, Fei Xia, Ed~H. Chi, Quoc~V Le, and Denny Zhou.
\newblock Chain of thought prompting elicits reasoning in large language models.
\newblock In Alice~H. Oh, Alekh Agarwal, Danielle Belgrave, and Kyunghyun Cho, editors, \emph{Advances in Neural Information Processing Systems}, 2022.
\newblock URL \url{https://openreview.net/forum?id=_VjQlMeSB_J}.

\bibitem[Wu et~al.(2024{\natexlab{a}})Wu, Peng, Du, Zheng, Liu, Wu, Ma, Li, Yang, Zhou, et~al.]{wu2024comparative}
Siwei Wu, Zhongyuan Peng, Xinrun Du, Tuney Zheng, Minghao Liu, Jialong Wu, Jiachen Ma, Yizhi Li, Jian Yang, Wangchunshu Zhou, et~al.
\newblock A comparative study on reasoning patterns of openai's o1 model.
\newblock \emph{arXiv preprint arXiv:2410.13639}, 2024{\natexlab{a}}.

\bibitem[Wu et~al.(2024{\natexlab{b}})Wu, Sun, Li, Welleck, and Yang]{wu2024inference}
Yangzhen Wu, Zhiqing Sun, Shanda Li, Sean Welleck, and Yiming Yang.
\newblock Inference scaling laws: An empirical analysis of compute-optimal inference for problem-solving with language models.
\newblock \emph{arXiv preprint arXiv:2408.00724}, 2024{\natexlab{b}}.

\bibitem[Yuksekgonul et~al.(2024)Yuksekgonul, Bianchi, Boen, Liu, Huang, Guestrin, and Zou]{yuksekgonul2024textgrad}
Mert Yuksekgonul, Federico Bianchi, Joseph Boen, Sheng Liu, Zhi Huang, Carlos Guestrin, and James Zou.
\newblock Textgrad: Automatic" differentiation" via text.
\newblock \emph{arXiv preprint arXiv:2406.07496}, 2024.

\end{thebibliography}
	\newpage
	\setcounter{section}{0}
\renewcommand{\thesection}{\Alph{section}}
\vspace{-2mm}\section{Methods}\vspace{-1mm}
\label{sec:method}
\subsection{Electronic Circuit Principles (\modelname{})}\vspace{-1mm}

\paragraph{Faraday's Law of Inference-Time Learning}
Inference-time learning (ITL), also named as in-context learning, can be likened to the behavior of a semantic magnetic field generated by a sub-power source, as described by Faraday's Law. The positive direction of the semantic magnetic field is defined by the query semantic vector $\overrightarrow{S}_q$.
Given $N$ context samples represented by semantic vectors $\overrightarrow{S}_i$, the total semantic magnetic field strength $\Phi^B_0$ experienced by the model is the projection of these vectors onto $\overrightarrow{S}_q$.  Mathematically:
\begin{equation}
    \Phi^B_0 = \sum^N_{i=1} \cos \theta_{qi} \cdot \left\lvert \overrightarrow{S}_i \right\rvert = \sum^N_{i=1} \frac{\overrightarrow{S}_q \cdot \overrightarrow{S}_i}{\left\lvert \overrightarrow{S}_q \right\rvert}\label{eq:phy-b}
\end{equation}
where $\cos \theta_{qi}$ represents the angle between $\overrightarrow{S}_q$ and $\overrightarrow{S}_i$.

As new context samples are processed, the semantic field strength $\Phi^B_0$ is hypothesized to decay linearly over time:
\begin{equation}
    \Phi^B(t) = -\lambda \Phi^B_0 t,
\end{equation}
where \(\lambda\) is the decay constant representing the rate at which the model forgets or absorbs information.
According to Faraday’s Law of electromagnetic induction, the rate of change in the semantic magnetic field induces an electromotive force, which drives effective ITL. With a time step \(\Delta t\) for each reasoning step, the induced EMF, \(\mathcal{E}_{\text{ITL}}\), is:
\begin{equation}
    \mathcal{E}_{\text{ITL}} = -\frac{\Delta \Phi^B(t)}{\Delta t} = \lambda \Phi^B_0.
\end{equation}
Substituting \(\Phi^B_0\) into this equation yields:
\begin{equation}
    \mathcal{E}_{\text{ITL}} = \lambda \sum^N_{i=1} \frac{\overrightarrow{S}_q \cdot \overrightarrow{S}_i}{\left\lvert \overrightarrow{S}_q \right\rvert}.
\end{equation}

In conclusion, this analogy offers a structured framework to analyze how information flows during ITL. By extending the dynamics of ITL through the principles of electromagnetic induction, we provide a mathematical principle to explain how models incorporate and gradually forget information as they process contextual examples.

\paragraph{Ohm's Law of Inference-Time Reasoning}
In this framework, each sub-difficulty of a model introduces a distinct level of difficulty, analogous to resistors in series.
Formally, following \citet{chen2024unlocking}, the ease with which a model can solve a particular task is represented by its reasoning boundary. Let \( \mathcal{B}_{\text{ITR}} \) denote the overall reasoning boundary of a Inference-Time Reasoning process, and \( \mathcal{B}_{i} \) denote the reasoning boundary of the \(i\)-th sub-capability. The overall reasoning boundary can be expressed by the harmonic sum:
\begin{equation}
\mathcal{B}_{\text{ITR}} = \frac{1}{\sum^K_{i=1} \frac{1}{\mathcal{B}_{i}}}, \label{eq:granularity}
\end{equation}
where $K$ denotes the number of sub-capabilities.
Reasoning resistor, $R$, quantifies the difficulty of a reasoning process and is mathematically defined as the reciprocal of the reasoning boundary. A higher boundary corresponds to an easier problem and lower resistance. Specifically, we define the sub-difficulty of the $i$-th sub-task as:
\begin{equation}
    R_i = \frac{1}{\mathcal{B}_{i}}.
\end{equation}
Substituting this into Equation~\ref{eq:granularity}, the total reasoning resistor for the ITR process is expressed as:
\begin{equation}
    R_{\text{ITR}} = \sum^K_{i=1} R_i, \label{eq:resistor}
\end{equation}
where $R_{\text{ITR}}$ represents the combined difficulty, analogous to a total resistor in a series circuit.
Overall, the ITR reasoning framework is further formalized by the Ohm's law analogy:
\begin{equation}
    I_{model} = \frac{\mathcal{E}_{\text{model}} + \mathcal{E}_{\text{ITL}}}{R_{\text{ITR}} + R_0},
    \label{eq:ohm2}
\end{equation}
where $R_0$ denotes an output resistor with static resistance value. This modular formulation effectively quantifies cognitive task complexity through a combined reasoning resistor.

\vspace{-2mm}\subsection{General Experimental Settings}\vspace{-1mm}
All experiments, unless stated otherwise, are conducted on GPT-3.5-Turbo. Following \citet{wei2022chain}, ITR experiments select 5 manually constructed demonstrations for all multi-step reasoning tasks. \textit{Top-p} is selected from $\{0.95, 1\}$, and \textit{Temperature} is selected in range \([0, 1]\) for robustness evaluation.
To formalize the demonstration selection process, we retrieve the semantic magnetic field strength \(\Phi^B_0\) by default, as defined in Eq.~\ref{eq:phy-b}.  In addition, only the magnetic field strength $\Phi^B_0$ is calculated and the resistor value $R_{\text{ITR}}$ directly adopts the fitting values from \citet{chen2024unlocking}. Further, other parameters are fitted values using the verification set. Different models fit different voltages $\mathcal{E}_{model}$. The whole experiment shares a $R_0$ value. Different demonstration representation methods have different $\lambda$. All experiments involving variance were sampled 3 times at different temperatures.
Notably, in addition to the validation experiments, we predominantly utilize the BigGSM~\citep{chen2024unlocking} and BigGSM++~\citep{chen2025rbf++} datasets, which precisely annotate reasoning difficulty across global planning, local operations, and calculation sub-tasks. In ITL, demonstrations are dynamically retrieved using a 5-shot approach from the GSM8K development dataset~\citep{cobbe2021training} by default.

\subsection{Verification Experiment Details}
In the effectiveness validation section of \modelname{}, and with the aim of comprehensively assessing the efficacy of our tasks, we first classify each task’s difficulty (hereafter ``resistance'') into four categories: global planning difficulty, local operational difficulty, domain‐knowledge analysis difficulty, and computational difficulty (the latter being relevant only to mathematical tasks). The total resistance is then defined as:
\begin{equation}
    R_{\text{ITR}} = R_{\text{plan}} + R_{\text{operation}} + R_{\text{domain}} + R_{\text{calculate}},
\end{equation}
where $R_{\text{plan}}$ denotes the resistance due to global planning difficulty, $R_{\text{operation}}$ denotes the resistance due to local operational difficulty, $R_{\text{domain}}$ denotes the resistance due to domain‐knowledge analysis difficulty, and $R_{\text{calculate}}$ denotes the resistance due to computational difficulty (set to zero for non‐computational tasks).

\begin{figure}[t]
    \centering
    \includegraphics[width=\textwidth]{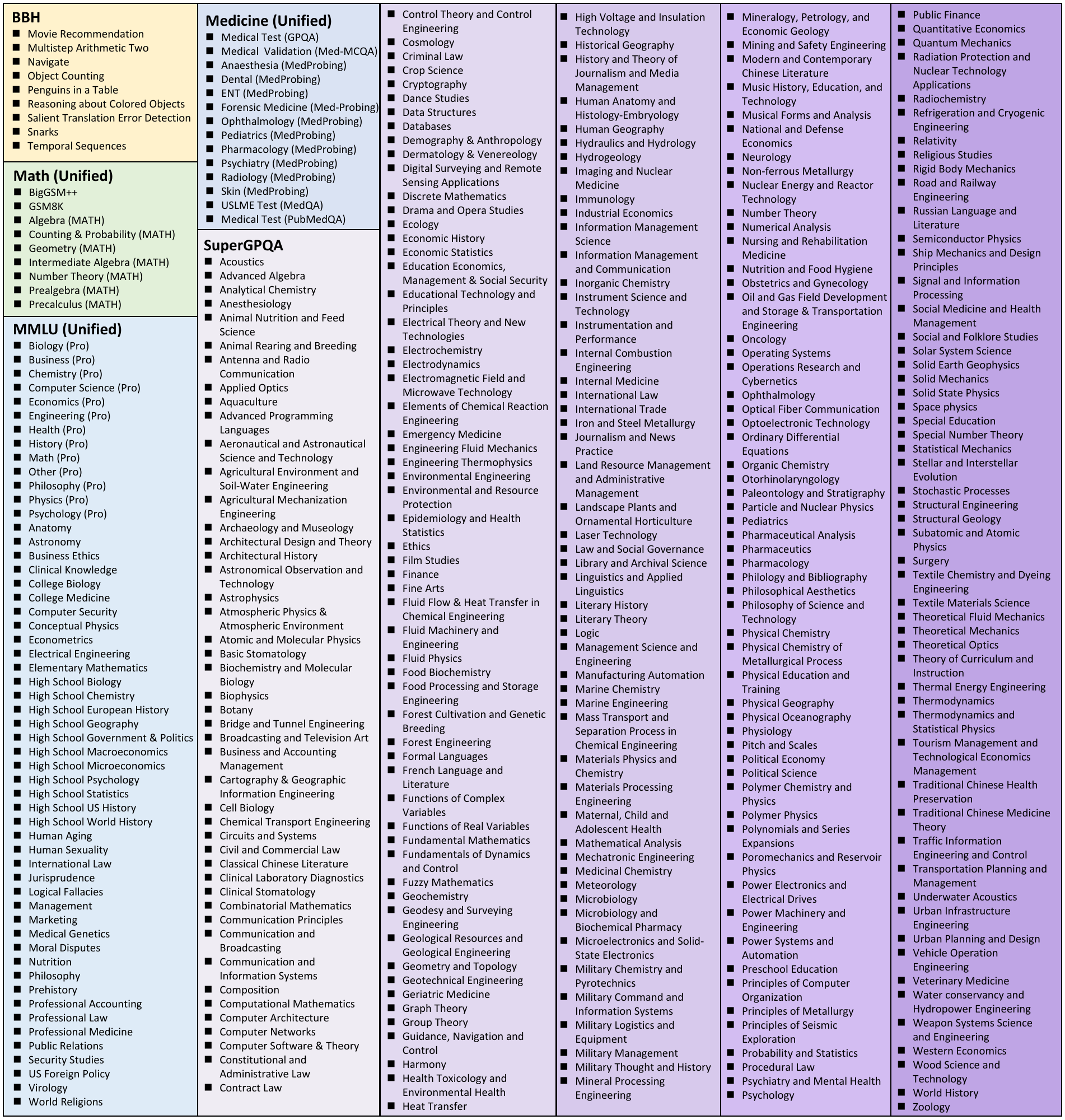}
    \caption{
      \textbf{Over 350 tasks in unified evaluation data.}
    }
    \label{fig:tasks}
    
\end{figure}

Furthermore, as illustrated in the Fig.~\ref{fig:tasks}, the task evaluation benchmarks are primarily drawn from the BigGSM~\citep{chen2024unlocking}, BigGSM++~\citep{chen2025rbf++}, MedProbing~\citep{cheng2024adapting}, Med-MCQA~\citep{pal2022medmcqa}, MedQA~\citep{jin2021disease}, PubMedQA~\citep{jin-etal-2019-pubmedqa}, BBH~\citep{suzgun-etal-2023-challenging}, MMLU~\citep{hendrycks2021measuring}, MMLU-Pro~\citep{wang2024mmlupro}, GPQA~\citep{rein2024gpqa}, and SuperGPQA~\citep{du2025supergpqa} datasets with over 350 tasks.

First, we prompt multiple LLMs to generate several reasoning trajectories, and then retain only those trajectories that produce correct answers. We develop corresponding automatic annotation scripts for each task, mathematical tasks typically employ a dedicated annotation script per task, whereas other STEM tasks use a universal annotation script. In practice, the number of “\textbackslash n\textbackslash n” separators or “Step x:” markers indicates the number of planning steps required and thus the overall planning difficulty; likewise, the number of clauses or bullet points reflects the local operation difficulty.

The annotation of the number of global and local reasoning steps required depends on the model-generated correct rationales and is automated through a manually constructed step-segmentation annotation program. Furthermore, since the granularity of the model’s reasoning does not fully align with that of humans, and the segmentation program cannot perform fine-grained annotations for every case, a considerable amount of annotation noise exists in the validation experiments across nearly 350 tasks, resulting in a relatively low $R^2$. However, under a strictly human-annotated setting for global and local step counts, such as that of the BigGSM++ dataset, the $R^2$ can exceed $0.9$. Therefore, except for the validation experiments, all subsequent experiments are conducted on the BigGSM++ dataset.

Inspired by \citet{chen2024unlocking}, we treat the domain difficulty as a constant for each task; this constant was obtained via parameter search using Newton’s method in a validation experiment on 10\% of the GPT-3.5-turbo generated dataset. For mathematical tasks, we design task-specific regular-expression extraction rules to estimate computational difficulty; however, since most operations not involving large numbers yield a computational difficulty close to zero, we omit this factor for some math tasks.

\paragraph{ITR Ohm's Law Verification}
To validate the Ohm's Law assumption in the ITR approach, we conduct the following experiment. To eliminate potential interference from ITL and ensure effective stimulation of ITR capabilities, we append the phrase ``Let's think step-by-step!" to the user query. This adjustment aims to stimulate and isolate the ITR mechanism and prevent ITL-related influences.
We then perform sampling based on power levels from the \textsc{BigGSM} dataset, using task-specific unit intervals to map calculated power to accuracy\footnote{Since the distribution of reasoning difficulty varies across tasks, a sufficiently large sample size is required to accurately compute accuracy. Accordingly, we employ different units for each task.}. To reduce random errors and ensure sufficient sampling points, we retain data points with at least 10 samples. Accuracy is calculated for each sampling interval to analyze the relationship between power levels and model performance, enabling an effective evaluation of the correlation between output power and accuracy to verify Ohm's Law. 

\begin{figure*}[t]
    \centering
    \includegraphics[width=\textwidth]{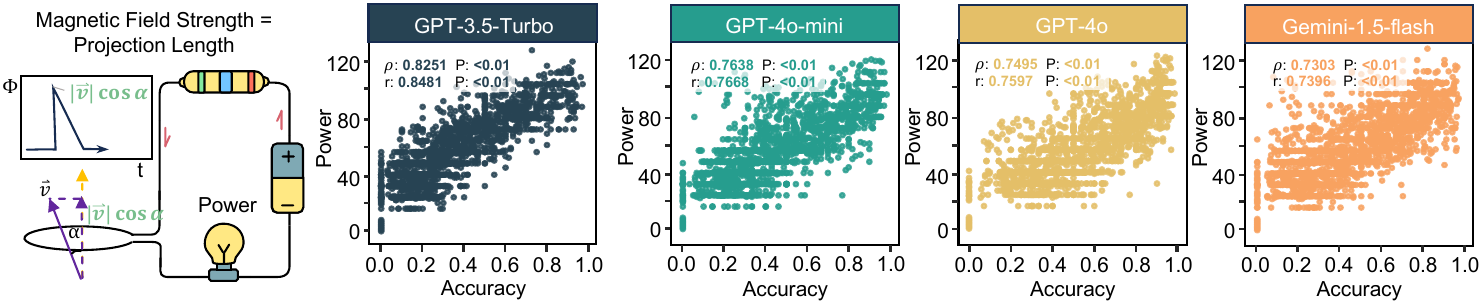}
    \caption{
      \textbf{The Analysis for ITL Faraday’s Law Verification Across 350 Tasks and 4 Models.}
    }
    \label{fig:itl-verification-append}
    
\end{figure*}

\paragraph{ITL Faraday's Law Verification}
To validate the semantic magnetic field hypothesis in ITL, we observe that without ITR reasoning, performance on complex mathematical tasks often declines to near-zero levels, thereby cannot easily observe their relevance. Thus, this experiment incorporates both ITR reasoning and ITL. While the ITR reasoning format remains constant, we focus on varying the similarity measures used for sample retrieval through ITL.
The central adjustment involves applying different similarity measures for sample retrieval, directly affecting the model's performance on complex tasks. These measures also help calculate the additional ``voltage" introduced by semantic magnetic fields, quantifying their influence on performance. All demonstration representation methods employ BGE~\cite{chen2024bge} to represent the request in the hidden space.

\vspace{-2mm}\subsection{Optimization Interpreting Experiments through Magnetic Field Adjustment}\vspace{-1mm}

\paragraph{Encoder Comparison}
To evaluate how different encoders influence model performance, we replace the sample encoder with BERT~\cite{devlin-etal-2019-bert}, RoBERTa~\cite{liu2019roberta}, and BGE~\cite{chen2024bge}. These embeddings are used for sample retrieval in ITL to minimize potential biases that may arise if a single encoder disproportionately emphasizes specific data features. For instance, an encoder might favor samples aligned with its encoding patterns, leading to distortions in the relationship between retrieved representations and model performance. By employing diverse encoders, we aim to mitigate such biases and promote more accurate theoretical modeling for ITL.\vspace{-5pt}

\paragraph{Meta-Recognition Representations} In this experiments, all demonstration representation methods employ the meta-recognition~\cite{didolkar2024metacognitive} framework by default. Specifically, GPT-4o~\cite{achiam2023gpt} is utilized to extract sub-capability tags, and then RoBERTa~\cite{liu2019roberta} concatenates and represents these tags in the hidden space.

\paragraph{Synthetic-CoT Implementation} 
This section examines the implementation of \textit{Synthetic-CoT} within the \modelname{} framework to assess its impact on reasoning performance. Synthetic-CoT autonomously generates in-context demonstrations with Inference-Time Reasoning reasoning paths. We hypothesize that these demonstrations align the system’s magnetic field, reducing logical resistance and enhancing overall circuit power output.
To validate this, we generate a series of demonstrations by \textit{Synthetic-CoT}. Specifically, In-context demonstrations generated by \textit{Synthetic-CoT} are positively aligned with the magnetic polarity of user queries direction.
We analyze the magnetic field contributions of these categories, focusing on their differing effects. Subsequently, we integrate the computed power and measured accuracy to further investigate their operational mechanisms.

\vspace{-2mm}\subsection{Optimization Interpreting Experiments through Resistor Reorganization}\vspace{-1mm}
\paragraph{Direct answer output}  
We analyze the effect of \textit{direct answer output} on reasoning performance through the \modelname{} framework. This approach bypasses intermediary reasoning steps by appending the directive, ``Let’s directly output the result without additional reasoning content,'' which generates a significant logical resistor within the system. To quantify this, we gradually improve the resistance value of reasoning resistor $R_{\text{ITR}}$, which meets the largest Spearman correlations in the validation set, and employ the $R_{\text{ITR}}$ to the test set.\vspace{-5pt}

\paragraph{Tool-Usage and PoT}  
We further investigate how \textit{Tool-Usage} and \textit{Program-of-Thought (PoT)} mitigate specific resistors in the reasoning process, focusing on their impact under the \modelname{} framework on computational and planning resistors:
\begin{itemize}[leftmargin=4ex]
    \item \textbf{Tool-Usage}: Tool-Usage aims to reduce \textit{computational resistor} by offloading complex calculations to external tools. Modeled as a zero-resistor computational circuit in our experiments, Tool-Usage enhances the Power-Accuracy Spearman correlation.
    \item \textbf{Program-of-Thought (PoT)}: PoT targets \textit{planning resistor} by employing structured programming logic to enable clearer and more efficient reasoning. This advantage stems from PoT's superior ability to reduce planning resistor, facilitating finer-grained logical alignment.\footnote{All implementation prompts come from \citet{chen2024unlocking}.}
\end{itemize} 
These findings confirm that while Tool-Usage primarily addresses the single-step computational resistor, PoT further excels in optimizing the overall planning resistor. Both methods significantly enhance performance within the \modelname{} framework, highlighting their complementary contributions to reducing resistance and improving reasoning capabilities.\vspace{-5pt}

\paragraph{Self-consistency and Coverage Design} 
This experiment evaluated the self-consistency strategy proposed by \citet{wang2023selfconsistency}, which employs multiple parallel reasoning paths and a majority-vote mechanism for the final decision. This strategy is hypothesized to enhance performance by reducing the system's resistance to diverse inputs, analogous to parallel resistors in a circuit, where the total resistance decreases compared to series connections. By applying this analogy, we demonstrate how parallel reasoning can lower cognitive load, improving accuracy and efficiency in decision-making.

To examine the influence of inference scaling on model performance, we investigated the relationship between the number of parallel samples, \( k \), and the metric like Accuracy or Pass@k, as described in the Inference Scaling Law. Specifically, Pass@k measures the accuracy of the Coverage method which obtains a correct result when at least one sampled prediction is accurate. We modeled the reasoning process as a resistive circuit, where each sample represents a parallel resistor. As \( k \) increases, the resistance of reasoning resistor approaches zero, reflecting a refined reasoning process and enhanced accuracy. Further, Coverage methods introduce zero verification resistance introduce much less overall resistance values, larger output power, and higher related performance. The experiment systematically varied \( k \) from 1 to 100 in increments of 5, recording the accuracy and Pass@k metric under each condition. We track individual reasoning path resistance, comparing these values with observed accuracy to validate the circuit analogy. The effectiveness of Self-consistency and Coverage methods can be mathematically proved based on the \modelname{} detailed in Theoretical Proof.

We annotated 100 instances, each comprising 32 outputs sampled from the model, with four annotators involved in the labeling. For each instance, the 32 outputs were presented to the annotators in sequence; they were asked to indicate whether each output contained text or logical content that was exactly duplicated from any previously shown example. Finally, for sample sizes of $2^n$ , we counted the number of outputs exhibiting no duplication and treated these counts as independent samples.

\vspace{-2mm}\subsection{Optimization Exploration Experiments through Magnetic Field Adjustment}\vspace{-1mm}

\paragraph{Diversity}
This study examines the differing impacts of semantic diversity on static and dynamic ITL retrieval. We map all samples into the hidden space using RoBERTa. In static ITL, we employ a fixed set of demonstrations with the highest semantic diversity across the task, hypothesizing that this diversity enhances model performance by providing a wider range of contextual cues. In contrast, dynamic ITL selects demonstrations in real-time based on the test sample’s characteristics, prioritizing semantically similar examples that align closely with the input.
To test these, we evaluate both retrieval strategies. For static ITL, we compare \modelname{} exposed to high-diversity and random demonstration sets, measuring performance using accuracy. For dynamic ITL, we contrast random selection with retrieval strategies emphasizing semantic alignment. We hypothesize that semantic diversity improves static ITL by stabilizing decision-making, while in dynamic ITL, retrieving highly relevant demonstrations yields better accuracy. In addition, in order to prove that diversity is ineffective in the setting of dynamic ITL, we sort the semantic magnetic field strength and select as diverse samples as possible among the top-$k$ samples as ITL demonstrations.\vspace{-5pt}

\paragraph{Demo-Optimization}
This experiment introduces demo-optimization, a novel method for optimizing demonstrations in ITL. The approach dynamically refines demonstrations based on feedback from the LLM's outputs, which serve as gradients for improvement. Specifically, initial ITL demonstrations in prompting are selected, with both development dataset for optimization and optimization prompt configurations prepared. A feedback loop guides adjustments to the demonstrations, improving their clarity, relevance, and alignment with prompt requirements. Demonstrations are iteratively refined, and evaluated for contextual alignment with user requests in the development dataset. The method is compared against static and whole-prompt-optimized baselines, with performance assessed using task accuracy, output consistency, and learning efficiency.

\vspace{-2mm}\subsection{Optimization Exploration Experiments through Resistor Reorganization}\vspace{-1mm}
\paragraph{ITR Optimization Proportion}
\begin{figure}[t]
    \centering
    \includegraphics[width=0.9\textwidth]{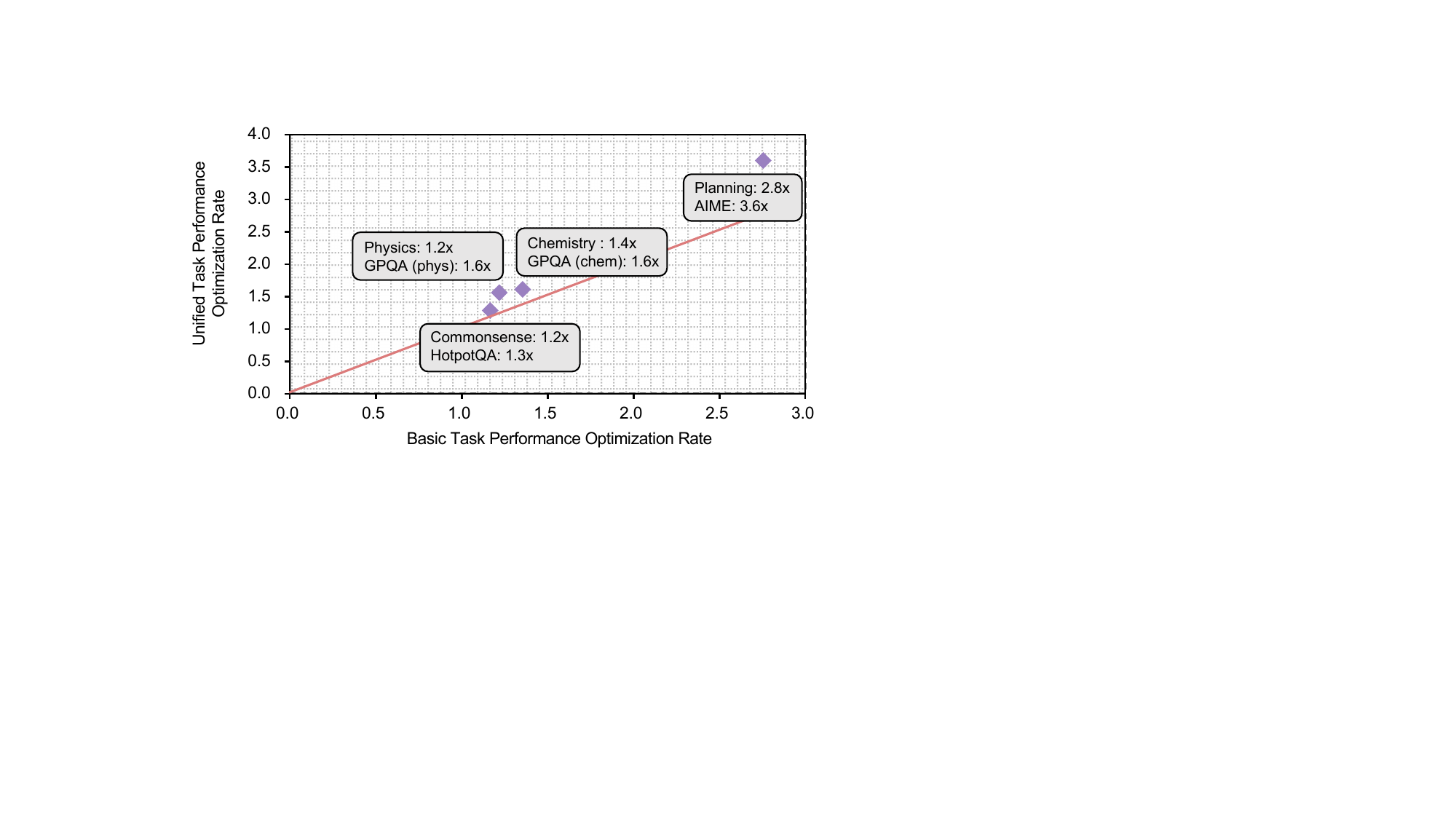}
    \caption{Meta-analysis for the model improvements on different tasks. The data comes from \citet{wu2024comparative,valmeekam2024llms} and \citet{openai2024o1}. We collect evaluation data related to plan and knowledge based on \citet{openai2024o1}. In addition, we remove data points with final results exceeding 90\%, because these data have exceeded human levels, resulting in potential labeling bias and test set labeling errors.}\label{fig:meta-analysis}
\end{figure}

To evaluate the improvement in reasoning resistance \( R_i \) to the target level of \( \frac{1}{k} \), which depends heavily on the task's reasoning demands, we conducted a meta-analysis of experimental results from diverse datasets. These datasets are divided into two main categories: (1) tasks requiring advanced planning and logical reasoning, such as mathematical derivations and multi-hop question answering, and (2) tasks reliant on extensive domain-specific knowledge, such as chemistry and physical question answering.
As shown in Fig.~\ref{fig:meta-analysis}, our meta-analysis demonstrates that tasks requiring advanced planning gain the most from enhanced exploratory reasoning, significantly improving challenging components (e.g., a 2.8-fold optimization in planning capability) and overall task performance (e.g., a 3.6-fold optimization on AIME, a benchmark emphasizing long-term planning). In contrast, tasks relying heavily on domain-specific knowledge exhibit limited improvement without targeted optimization in those areas. This underscores the importance of domain-aware reasoning strategies. For example, tasks involving complex reasoning and knowledge integration, such as GPQA (chemistry, physics) and HotpotQA, show performance improvements directly proportional to advancements in the related knowledge domains (approximately 1:1 ratio). The effectiveness and the underlying reasons for such phenomena can be mathematically analyzed and proved based on the \modelname{} detailed in Theoretical Proof.\vspace{-5pt}

\paragraph{Fine-grained Self-consistency}

In this experiment, we examine how decomposing a self-consistency ensemble into smaller, incremental majority-voting steps reduces the reasoning resistance in circuits, compared to the conventional single-step approach. The circuit operates at a stable supply voltage of 5 units, and reasoning resistance value and output power are evaluated as the number of steps increases. Each majority-voting step simplifies logical complexity and reduces overall resistance by incorporating less verification resistance in the decision-making process. Our findings demonstrate a consistent decline in reasoning resistance alongside a rise in output power, confirming that finer-grained majority-voting processes improve both efficiency and power output in circuits. The effectiveness of this method can be mathematically proved based on the \modelname{} in Theoretical Proof.\vspace{-5pt}

\paragraph{Chain-of-Verification}
Traditional self-consistency frameworks rely on a sequential verification, where each step depends on the completion of the previous one. This structure often incurs high computational costs, as subsequent tasks cannot begin until earlier ones are resolved. To overcome this limitation, we introduce the ``chain-of-verification (CoV)," a novel strategy that parallelizes verification tasks into independent components. Each component operates concurrently, validating distinct parts of the system without interdependence. This parallel approach reduces overall computational resistance by distributing the workload across multiple verifiers, thereby enabling more effective validation. As illustrated in Fig.~\ref{fig:ITR-optimization}\textit{b}, parallelization not only accelerates the verification process but also enhances model performance by mitigating bottlenecks. This improvement is particularly pronounced in large-scale systems, where sequential verification would otherwise impede efficiency. The effectiveness of CoV can also be mathematically proved in Theoretical Proof.

\vspace{-2mm}\subsection{Best Practice}\vspace{-1mm}
In practice, we integrate all proposed strategies into the o1-preview model to address IOI, IMO, and idea-generation tasks for much less resistance value of planning resistor.\vspace{-5pt}

\paragraph{IOI Task}
For the IOI task, we employ the following methods: First, using o1-preview, we maximize the optimization of the resistance value of its ``planning resistor." Additionally, five high-intensity semantic field problems from the code\_contest training dataset are retrieved as demonstrations, boosting the voltage supplied by additional ITL fields. Second, we apply the Coverage technique, sampling 100 instances and leveraging both provided and model-generated test cases in parallel, effectively reducing resistance in Inference-Time Reasoning reasoning. This process, executed in parallel, effectively reduces the resistance in the Inference-Time Reasoning reasoning. Furthermore, we incorporate the CoV concept to iteratively validate the final outputs, thereby optimizing the final performance. Additionally, the CoV framework is employed to iteratively validate outputs and optimize overall performance. To prevent data leakage from competition datasets in C++, competition code is generated in Python. Manual adjustments ensure compatibility between Python and C++ input/output handling, preserving the core logic for fair comparative testing.\vspace{-5pt}

\paragraph{IMO Task}
A similar approach is adopted for the IMO task. o1-preview is optimized for planning resistor, and five high-intensity semantic field problems from past IMO datasets are retrieved as demonstrations to enhance ITL field voltage. The Coverage technique is again employed, sampling 100 instances in parallel to reduce reasoning resistance. We introduce the fine-grained self-consistency approach within the CoV framework, enabling iterative validation of outputs and achieving notable performance gains. Additionally, we only evaluate answers and do not consider process scores.\vspace{-5pt}

\paragraph{Idea-Generation Task}
We utilize the AI Scientist framework~\citep{lu2024aiscientist} to generate over 10 ideas per domain and draft academic papers, limited to the abstract and introduction sections to address ethical concerns by avoiding experimental claims. A 16-member review panel, including 2 Superior reviewers (6+ years of experience), 8 Senior reviewers (4+ years), and 6 Junior reviewers (2+ years), ensures reliability. Each paper is reviewed by at least four reviewers, including two Senior or Superior reviewers, following NeurIPS guidelines. High-quality examples with strong semantic fields, sourced from past ITLR, NeurIPS, and CoLM openreview data, are used to guide generation. The Coverage technique, sampling of 100 instances, and iterative validation through the CoV framework further refine the process.

\begin{figure}[t]
    \centering
    \includegraphics[width=\textwidth]{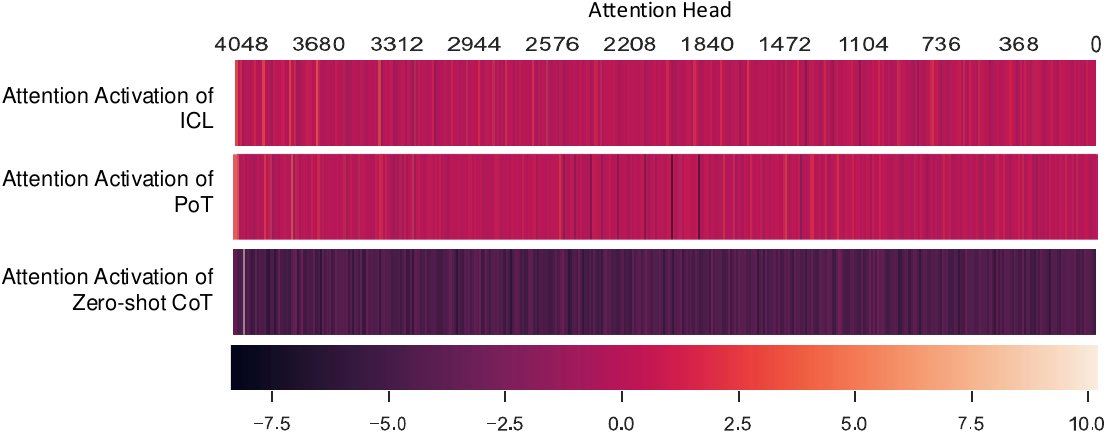}
    \caption{Neuron activation heatmaps for the most activated attention layer under three reasoning strategies: (a) Zero-shot ITR, (b) ITL with Semantic Magnetic Field Retrieval, and (c) Program of Thought (PoT).}\label{fig:explanation}
\end{figure}
\subsection{Internal Mechanism underlying the \modelname{}}
To investigate the internal mechanisms underlying \modelname{}, we conduct experiments  in the LLaMA3-8B~\cite{dubey2024llama} focusing on three strategies: Zero-shot ITR, ITL with Semantic Magnetic Field Retrieval, and PoT. By analyzing neuron activation values and frequencies, particularly in the most activated attention layer, we aim to elucidate the relationship between reasoning strategies and neural dynamics. These findings support the theoretical framework of \modelname{}.
The experiments use the BigGSM dataset, a standard benchmark for mathematical reasoning tasks. For ITL, demonstrations are dynamically retrieved using the BGE model in a 3-shot configuration.
Fig.~\ref{fig:explanation} shows neuron activation heatmaps for the most activated attention layer under each reasoning strategy. For Zero-shot ITR, the heatmap reveals sparse activations, with limited regions of intense activity, highlighting the difficulty of reasoning without contextual support. Conversely, ITL and PoT display more distributed and elevated activation patterns. These suggest that the underlying reasoning mechanisms engage a broader spectrum of neurons, with comparable activation dynamics reflecting similar reasoning power across these strategies.

\section{Theoretical Proof}
\subsection{Analysis of Effectiveness of Self-consistency and Coverage Design}
\label{append:proof-self}
\subsubsection{Analysis of Effectiveness of Self-consistency}
First, according to the series resistor assumption of ITR, the total resistance of series resistors can be expressed as:
\begin{equation}
    R_{\text{all}} = R_0 + R_{\text{ITR}}.
\end{equation}
Building upon the illustration in Fig.~\ref{fig:ITR-strategy}c, we hypothesize that self-consistency, achieved through multiple parallel reasoning processes and subsequent result aggregation by majority voting, can be represented as an aggregation resistor integrated alongside several parallel ITR resistors. In this framework, the effective resistance of the reasoning process transitions from  $R_{\text{ITR}}$ to $\frac{1}{\sum_{i=1}^n\frac{1}{R_{\text{ITR}}^i}}$, accompanied with a majority-voting resistor $R_S$ to determine the final outcome. Consequently, the total resistance associated with self-consistency, denoted as $R^{'}_{\text{all}}$, is expressed as:
\begin{equation}
    R^{'}_{\text{all}} = R_0 + R_S + \frac{1}{\sum_{i=1}^n\frac{1}{R_{\text{ITR}}^i}},\label{eq:r-all-self}
\end{equation}
where $n$ is the number of samples, \(R_{\text{ITR}}^i\) denotes the resistance corresponding to the \(i-\)th sample and $R_S$ accounts for the resistor due to self-consistency integration.
This framework allows for rapid computation of the output power:
\begin{equation}
    P^{\text{self}}_{\text{out}} = \frac{\left(\mathcal{E}_{\text{model}} + \mathcal{E}_{\text{ITL}}\right)^2 R_0}{(R_0+R_S + \frac{1}{\sum_{i=1}^n\frac{1}{R_{\text{ITR}}^i}})^2}.
\end{equation}
As the number of samples increases, the curves in Fig.~\ref{fig:ITR-strategy}c reveal that the real performance curve is almost exactly consistent with the theoretical range.
Furthermore, assume that \(\{R_{\text{ITR}}^i\}_{i=1}^\infty\) is bounded thus we can easily prove that the overall resistance has a lower bound:
\begin{equation}
    \lim_{n\rightarrow +\infty}R'_{\text{all}} = R_0 + R_S + \lim_{n\rightarrow +\infty}\frac{1}{\sum_{i=1}^n\frac{1}{R_{\text{ITR}}^i}} = R_0 + R_S.
\end{equation}
The corresponding performance bound is:
\begin{equation}
    \hat{P}^{\text{self}}_{\text{out}} = \frac{\left(\mathcal{E}_{\text{model}} + \mathcal{E}_{\text{ITL}}\right)^2 R_0}{(R_0+R_S)^2}
\end{equation}
As shown in Fig.~\ref{fig:ITR-strategy}c, model performance aligns with this theoretical upper bound as sample size increases.

\subsubsection{Analysis of Effectiveness of Coverage Design}
Similarly, we hypothesize that coverage can be considered correct if at least one of multiple parallel reasoning is accurate. Under this assumption, the integration difficulty in self-consistency verification approaches zero. Analogously, this can be modeled as a zero-value combined resistance in parallel with multiple ITR resistors, where the coverage resistance of self-consistency $\hat{R}_{\text{all}}$ is given by:
\begin{equation}
    \hat{R}_{\text{all}} = R_0 + R_S + \frac{1}{\sum_{i=1}^n\frac{1}{R_{\text{ITR}}^i}}= R_0 + \frac{1}{\sum_{i=1}^n\frac{1}{R_{\text{ITR}}^i}}
\end{equation}
where $n$ denotes the number of samples. Using this, the theoretical output power can be similarly calculated as:
\begin{equation}
    P^{\text{cover}}_{\text{out}} = \frac{\left(\mathcal{E}_{\text{model}} + \mathcal{E}_{\text{ITL}}\right)^2 R_0}{(\frac{1}{\sum_n\frac{1}{R_{\text{ITR}}}} + R_0)^2},\label{eq:p-cover}
\end{equation}
which are much more smaller than $P^{\text{self}}_{\text{out}}$ in the same $n$ sample size with $R_S \gg 0$. 
As the number of samples increases, the curve in Fig.~\ref{fig:ITR-strategy}c demonstrates that the real performance aligns closely with the theoretical range. Building upon the same bounded assumption of \(\{R_{\text{ITR}}^i\}_{i=1}^\infty\), this setup also defines a resistance lower bound:
\begin{equation}
    \lim_{n\rightarrow +\infty}\hat{R}_{\text{all}} = R_0 + \lim_{n\rightarrow +\infty}\frac{1}{\sum_{i=1}^n\frac{1}{R_{\text{ITR}}^i}} = R_0
\end{equation}
and the corresponding lower-bound performance:
\begin{equation}
    \hat{P}^{\text{cover}}_{\text{out}} = \frac{\left(\mathcal{E}_{\text{model}} + \mathcal{E}_{\text{ITL}}\right)^2}{R_0}.
\end{equation}
As shown in Fig.~\ref{fig:ITR-strategy}c, the model's performance gradually converges toward this upper bound and much larger than the performance of self-consistency, demonstrating consistency with the theoretical predictions.

\paragraph{NOTE}
According to the theoretical formulation in Eq.~\ref{eq:p-cover}, in the reasoning phase where the resistance is significantly greater than \( R_0 \), the overall trend shows that Coverage is approximately proportional to the number of samples. Meanwhile, certain Inference Scaling Laws indicate that Coverage and the logarithm of the number of samples exhibit an approximately proportional relationship. These two observations are not contradictory, as \modelname{} assumes relatively independent sampling. However, we have observed that when the number of samples exceeds 100, an increasing proportion of the samples become duplicates rather than independent samples. Consequently, the probability of generating new samples follows the trend where \( n \) samples are expected to yield approximately \( \log n \) distinct reasoning samples.

\subsection{Analysis of Effectiveness of Fine-grained Self-consistency}\label{append:proof-fine}
According to Eq.~\ref{eq:r-all-self}, the original circuit can be considered as a large parallel reasoning and verification process and the total resistor is as follows:
\begin{equation}
    R'_{\text{all}} = R_0 + R_S + \frac{1}{\sum_{i=1}^n\frac{1}{R_{\text{ITR}}^i}}.
\end{equation}
As shown in Fig.~\ref{fig:ITR-optimization}a, this process is decomposed into multiple sequential, smaller-scale parallel reasoning and verifications. This refinement transforms \textit{a single large parallel reasoning resistance with significant verification resistance} into \textit{a sequence of smaller parallel reasoning resistance and smaller verification resistance}. Formally, this is expressed as:
\begin{equation}
    R^{\text{fine}}_{\text{all}} = R_0 + \sum_j R_j^S + \sum_j\frac{1}{\sum_{i=1}^n\frac{1}{R_j^{i}}},
\end{equation}
where $n$ denotes the sampling size, $ R_j^S $ represents the small and fine-grained verification resistance for $j$-th step, and $ R_j^{i} $ corresponds to smaller reasoning resistance for $j$-th step for $i$-th sampling. The total resistance can be broken down into resistance of sub-step resistor as $ R^i_{\text{ITR}} = \sum_j R_j^{i} $. 

We aim to demonstrate that our approach indeed optimizes the total resistance, i.e., $ R^{\text{fine}}_{\text{all}} < R'_{\text{all}} $. To prove this, we need to show that $ R^{\text{fine}}_{\text{all}} - R'_{\text{all}} < 0 $. The proof is as follows:
\begin{align}
    R^{\text{fine}}_{\text{all}} - R'_{\text{all}} &= \sum_j R_j^S - R_S + \frac{R_{\text{ITR}}}{n} - \sum_j\frac{R_j^{i}}{n}\\
    &= \sum_j R_j^S - R_S.
\end{align}
It is evident that verifying correctness at the individual step level is simpler than verifying the correctness of the entire chain, so $ \sum_i R_i^S < R_S $. Therefore, $ R^{\text{fine}}_{\text{all}} - R'_{\text{all}} < 0 $, indicating that the total resistance of the circuit has been optimized.

\subsection{Analysis of Effectiveness of Chain-of-Self-consistency}\label{append:proof-chain}
Furthermore, we optimize the total circuit resistance from a different perspective, specifically by improving the resistance of validation resistor through self-consistency. In this approach, the validation resistor is decomposed into a parallel combination of multiple validation resistors and a meta-validation resistor with resistance $R_{\text{meta}}$, which verifies the correctness of the validation process. The original equation is thus reformulated as:
\begin{align*}
R^{\text{CoV}}_{\text{all}} &= R_0 +\frac{1}{\sum_{j=1}^k\sum_{i=1}^n \frac{1}{R_{\text{S}}}} + \sum_{j=1}^k\left(R_{\text{meta}}\right) + \frac{R_{\text{ITR}}}{n}\\
&= R_0 + \frac{R_{\text{S}}}{nk} + k R_{\text{meta}} + \frac{R_{\text{ITR}}}{n}.
\end{align*}
Clearly, as the sample size $n$ and the meta-validation steps $k$ increase, $\frac{R_{\text{S}}}{nk}$ decreases, leading to reduced difficulty in meta-validation. This significantly optimizes the total resistance and improves output power. Fig.~\ref{fig:ITR-optimization}b illustrates this performance enhancement.

However, as the number of samples requiring validation grows, the model's accuracy first increases and then decreases. Specifically, for sufficiently large validation steps, the resistance satisfies:
\begin{equation}
    \lim_{nk \to +\infty} R^{\text{CoV}}_{\text{all}} = R_0 + k R_{\text{meta}}.
\end{equation}
With excessive validation steps, $k$ becomes too large, resulting in $k R_{\text{meta}} > R_S$. Consequently, $R^{\text{CoV}}_{\text{all}}$ exceeds its optimized value, leading to a performance trend characterized by initial improvement followed by decline.

\subsection{Proof of Impact of Resistor Optimization}\label{append:proof-resistor}
To compute the derivatives of $ P_{\text{model}} $ with respect to $ R_{\text{ITR}} $, we start with the given expression:
\begin{equation}
P_{\text{model}} = \frac{\left(\mathcal{E}_{\text{model}} + \mathcal{E}_{\text{ITL}}\right)^2 R_0}{(R_{\text{ITR}} + R_0)^2}.
\end{equation}
According to Eq.~\ref{eq:resistor}, let $ R_{\text{ITR}} = R_1 + R_2 $, where $R_1 \gg R_2$, if $R_1$ is improved to $\frac{R_1}{k}$, the power can be calculated as:
\begin{equation}
    P^1_{\text{model}} = \frac{\left(\mathcal{E}_{\text{model}} + \mathcal{E}_{\text{ITL}}\right)^2 R_0}{(\frac{R_1}{k}+R_2 + R_0)^2}
\end{equation}
The corresponding power improvement is:
\begin{align}
    \Delta P^1_{\text{model}} &=P^1_{\text{model}}-P_{\text{model}} \\
    &= \frac{\left(\mathcal{E}_{\text{model}} + \mathcal{E}_{\text{ITL}}\right)^2 R_0}{(R_1+R_2 + R_0)^2}-\frac{\left(\mathcal{E}_{\text{model}} + \mathcal{E}_{\text{ITL}}\right)^2 R_0}{(\frac{R_1}{k}+R_2 + R_0)^2}\\
    &= \left(\mathcal{E}_{\text{model}} + \mathcal{E}_{\text{ITL}}\right)^2 R_0 (\frac{1}{(R_1+R_2 + R_0)^2}-\frac{1}{(\frac{R_1}{k}+R_2 + R_0)^2})
\end{align}

Similarly, if $R_2$ is improved to the original $\frac{R_2}{k}$, the power improvement can be calculated as:
\begin{equation}
    \Delta P^2_{\text{model}} = \left(\mathcal{E}_{\text{model}} + \mathcal{E}_{\text{ITL}}\right)^2 R_0 (\frac{1}{(R_1+R_2 + R_0)^2}-\frac{1}{(R_1+\frac{R_2}{k} + R_0)^2})
\end{equation}
To determine which improvement has a greater effect, consider:
\begin{equation}
    \Delta P^1_{\text{model}} - \Delta P^2_{\text{model}} = \left(\mathcal{E}_{\text{model}} + \mathcal{E}_{\text{ITL}}\right)^2 R_0 (\frac{1}{(\frac{R_1}{k} +R_2 + R_0)^2}-\frac{1}{(R_1+\frac{R_2}{k} + R_0)^2})
\end{equation}
Define: 
\begin{equation}
A = \frac{R_1}{k} + R_2 + R_0, \quad B = R_1 + \frac{R_2}{k} + R_0
\end{equation}
The fractional difference simplifies as: 
\begin{equation}
\frac{1}{A^2} - \frac{1}{B^2} = \frac{B^2 - A^2}{A^2 B^2}
\end{equation}
We need to prove that for \( R_1 \gg R_2 \) whether $B^2 - A^2 $ larger or smaller than zero. Now, let's calculate \( B^2 - A^2 \):
\begin{equation}
B^2 - A^2 = (B - A)(B + A)
\end{equation}
where: 
\begin{align}
B - A &= \left(R_1 + \frac{R_2}{k} + R_0\right) - \left(\frac{R_1}{k} + R_2 + R_0\right)\\
&= R_1 - \frac{R_1}{k} + \frac{R_2}{k} - R_2\\
&= R_1\left(1 - \frac{1}{k}\right) + R_2\left(\frac{1}{k} - 1\right)
\end{align}

and:
\begin{align}
B + A &= \left(R_1 + \frac{R_2}{k} + R_0\right) + \left(\frac{R_1}{k} + R_2 + R_0\right)\\
&= R_1 + \frac{R_1}{k} + R_2 + \frac{R_2}{k} + 2R_0
\end{align}

Analysis of the Sign of \( B^2 - A^2 \):
\begin{equation}
B^2 - A^2 = (B - A)(B + A)
\end{equation}
We only need to examine the sign of \( B - A \), since \( B + A > 0 \) holds true at all times.since \( R_1 \gg R_2 \), and \( 1 - \frac{1}{k} > 0 \) for \( n > 1 \), we have:
\begin{align}
R_1\left(1 - \frac{1}{k}\right) - R_2\left(1 - \frac{1}{k}\right) &>0\\
(R_1-R_2)\left(1 - \frac{1}{k}\right)&> 0\\
B^2 - A^2 &> 0\\
\Delta P^1_{\text{model}} - \Delta P^2_{\text{model}} &> 0
\end{align}
Thus, $\Delta P^1_{\text{model}} < \Delta P^2_{\text{model}}$ the difference is negative, and the proof is complete. Therefore, The greater the original difficulty of the task corresponding to the capability, the greater the impact on the final performance.

\end{document}